\documentclass[5p,times,twocolumn]{elsarticle}

\usepackage{amssymb}
\usepackage{amsmath}
\usepackage{xcolor}

\usepackage[caption=false,font=normalsize,labelfont=sf,textfont=sf]{subfig}

\usepackage{times}  
\usepackage{helvet}  
\usepackage{courier}  
\usepackage[hyphens]{url}  
\usepackage{graphicx} 
\usepackage{caption} 
\usepackage{cite}
\usepackage{algorithm}
\usepackage{amsfonts}
\usepackage{float}
\usepackage{algpseudocode}
\usepackage{multirow}
\usepackage{colortbl}

\usepackage{array} 
\newlength\savewidth\newcommand\shline{\noalign{\global\savewidth\arrayrulewidth
  \global\arrayrulewidth 1pt}\hline\noalign{\global\arrayrulewidth\savewidth}}
\newcommand{\tablestyle}[2]{\setlength{\tabcolsep}{#1}\renewcommand{\arraystretch}{#2}\centering\footnotesize}
\renewcommand{\paragraph}[1]{\vspace{1.25mm}\noindent\textbf{#1}}

\newcolumntype{x}[1]{>{\centering\arraybackslash}p{#1pt}}
\newcolumntype{y}[1]{>{\raggedright\arraybackslash}p{#1pt}}
\newcolumntype{z}[1]{>{\raggedleft\arraybackslash}p{#1pt}}

\newcommand{\app}{\raise.17ex\hbox{$\scriptstyle\sim$}}

\definecolor{deemph}{gray}{0.6}

\definecolor{baselinecolor}{gray}{.9}

\definecolor{dt}{HTML}{ADCAD8}
\definecolor{dt2}{HTML}{cddfe7}

\definecolor{defaultcolor}{HTML}{D3D3D3}
\newcommand{\default}[1]{\cellcolor{defaultcolor}{#1}}

\let\cite\citep

\definecolor{Gray}{rgb}{0.9,0.9,0.9} 
\usepackage{enumitem}
\usepackage{booktabs}
\usepackage{newfloat}
\usepackage{listings}
\usepackage{hyperref}

\begin{document}

\begin{frontmatter}

\title{MaMe \& MaRe: Matrix-Based Token Merging and Restoration \\ for Efficient Visual Perception and Synthesis}

\author[1]{Simin Huo}
\author[1]{Ning Li\corref{corres}}

\fntext[corres]{Corresponding Author: Ning Li (ning\_li@sjtu.edu.cn)}

\affiliation{
organization={Shanghai Jiao Tong University},
addressline={},
city={Shanghai},
postcode={200240},
state={Shanghai},
country={China}}

\begin{abstract}
Token compression is crucial for mitigating the quadratic complexity of self-attention mechanisms in Vision Transformers (ViTs), which often involve numerous input tokens. Existing methods, such as ToMe, rely on GPU-inefficient operations (e.g., sorting, scattered writes), introducing overheads that limit their effectiveness. We introduce MaMe, a training-free, differentiable token merging method based entirely on matrix operations, which is GPU-friendly to accelerate ViTs. Additionally, we present MaRe, its inverse operation, for token restoration, forming a MaMe+MaRe pipeline for image synthesis. When applied to pre-trained models, MaMe doubles ViT-B throughput with a 2\% accuracy drop. Notably, fine-tuning the last layer with MaMe boosts ViT-B accuracy by 1.0\% at 1.1$\times$ speed. In SigLIP2-B@512 zero-shot classification, MaMe provides 1.3× acceleration with negligible performance degradation. In video tasks, MaMe accelerates VideoMAE-L by 48.5\% on Kinetics-400 with only a 0.84\% accuracy loss. Furthermore, MaMe achieves simultaneous improvements in both performance and speed on some tasks. In image synthesis, the MaMe+MaRe pipeline enhances quality while reducing Stable Diffusion v2.1 generation latency by 31\%. Collectively, these results demonstrate MaMe's and MaRe's effectiveness in accelerating vision models. The code is available at \href{https://github.com/cominder/mame}{https://github.com/cominder/mame.}   
\end{abstract}







\end{frontmatter}

\section{Introduction}
\label{sec:intro}

Vision Transformers (ViTs)\cite{dosovitskiy2020image} have revolutionized computer vision by adopting the transformer architecture from natural language models\cite{vaswani2017attention}. However, the complexity of self-attention is quadratic $\mathcal{O}(N^2)$, where N represents the number of tokens. For applications requiring dense token representations, such as high-resolution images, this quadratic complexity presents a significant challenge, limiting the deployment of large-scale ViT models on resource-limited devices or in real-time applications.

To address the $\mathcal{O}(N^2)$ computational challenge, a straightforward yet effective approach is to reduce the number of tokens $N$ involved in the process. The strategies that have emerged include token pruning, token merging, and hybrid methods that integrate both. Pioneering works like DynamicViT\cite{rao2021dynamicvit} introduced a dynamic token sparsification framework that uses a lightweight, learnable prediction module to hierarchically prune tokens at various stages of the network. EViT\cite{liang2022evit} uses the class token to evaluate token importance, keeping the most attentive tokens while merging the others. Pruning's main drawback is irreversible information loss. Token merging combines similar tokens instead of discarding them. ToMe\cite{bolya2023tome} introduced a training-free method, using a fast bipartite soft matching algorithm to progressively merge similar tokens. Token Pooling\cite{marin2023tokenpooling} uses cluster analysis to aggregate information from neighboring tokens. DiffRate\cite{yu2023diffrate} makes compression rate differentiable to learn layer-wise rates, while Token Transforming\cite{zeng2025tokentransforming} generalizes both pruning and merging as specific cases of a broader matrix transformation, enabling more flexible, many-to-many mappings that can better preserve information. PruMerge \cite{shang2025prumerge} and VisionZip \cite{yang2025visionzip} first prune inattentive visual tokens based on attention scores and subsequently merge the remaining redundant tokens using spatial or semantic similarity matching for MLLMs.

Despite recent advancements, existing token reduction methods face several challenges. A primary issue is the \textbf{non-differentiable} nature of the token selection process when using the Top-K operation, which often requires complex workarounds for end-to-end training. Some methods are slow due to their reliance on clustering techniques like k-means, which are \textbf{computationally intensive} in practice. Additionally, many methods introduce \textbf{extra learnable parameters} for token selection or merging modules, leading to increased model complexity and training overhead. Lastly, an issue is the \textbf{dependency on specific architectures}; for example, EViT's reliance on a class token restricting its use in models where a class token might not be available.

To simultaneously address these limitations, inspired by ToMe, we introduce a training-free token merging approach that overcomes the mentioned challenges through several ways: 

\textbf{Differentiable Design}: Our method employs only differentiable operations throughout the token merging process, enabling seamless end-to-end training. By avoiding discrete operations, we maintain gradient flow and allow the model to be trained from scratch.

\textbf{Efficient Matrix Operations}: Instead of relying on operations such as clustering algorithms, sorting or explicit maximum selection,we utilize efficient, GPU-friendly full-matrix operations. This approach offers both theoretical efficiency and practical speedup.

\textbf{Parameter-Free Architecture}: Our approach introduces no additional learnable parameters, maintaining the original model's parameter, simplifying deployment, and reducing the complexity of model management.

\textbf{Plug-and-Play Integration}: Our approach can be directly applied to pre-trained models without any extra training, or seamlessly integrated during training from scratch. This flexibility significantly lowers the barrier to adoption.

Our contributions can be summarized as follows:

\begin{itemize}
\item We propose MaMe, a differentiable, and training-free token merging algorithm. MaMe leverages efficient full-matrix operations based on normalized cosine similarity to merge similar tokens while preserving distinct and critical tokens.

\item We introduce MaRe, the inverse operation of MaMe for token restoration, also realized through efficient full-matrix operations. The MaMe+MaRe pipeline can enhance image quality and reduce image synthesis latency.

\item We conduct extensive experiments across various tasks, including image classification, zero-shot classification, COCO Caption, video recognition and image generation. These demonstrate  MaMe's and MaRe's effectiveness.

\item We discuss how MaMe's design inherently preserves causality, making it a feasible candidate for efficiently reducing KV cache size in Large Language Models (LLMs).
\end{itemize}

This is an extended version of our paper titled "MaMe: Matrix-Based Token Merging" accepted to CVPR 2026 Findings.

\section{Related Work}
\label{sec:related}
To tackle the quadratic computational complexity caused by the self-attention mechanism in Vision Transformer models, researchers have proposed various methods to reduce the number of tokens fed into attention.

\subsubsection{Token Pruning}
Pruning methods hierarchically discard tokens deemed non-informative based on importance metrics. DynamicViT\cite{rao2021dynamicvit} pioneered this approach by attaching lightweight prediction heads at intermediate layers to score token relevance, using differentiable attention masking to enable end-to-end training. EViT\cite{liang2022evit} enhanced this framework by fusing pruned tokens into the class token, preserving partial information while reducing sequence length. Recent advancements include AdaViT\cite{meng2022adavit}, which extends pruning beyond tokens to attention heads and transformer blocks, implementing instance-adaptive computation graphs that allocate more resources to complex inputs. Recently, FastV\cite{chen2024fastv} revealed that visual token attention becomes highly sparse in the early layers of Multimodal Large Language Models (MLLMs), allowing for aggressive pruning of over 50\% of visual tokens without performance degradation. Similarly, VTW \cite{lin2025vtw} demonstrates that visual tokens can be entirely withdrawn after specific LLM layers, drastically improving decoding efficiency.However, these methods face fundamental limitations: 1) Early pruning decisions risk irreversible information loss, 2) Discrete selection operations create optimization challenges, and 3) Task-specific tuning is required for optimal threshold calibration. 

\subsubsection{Token Merging}
Merging techniques combine similar tokens rather than discarding them, preserving information while reducing computational load. ToMe\cite{bolya2023tome} revolutionized this area with training-free bipartite soft matching to merge the most similar token pairs at each layer. However, ToMe's fixed merge ratio per layer limits adaptability to varying input complexities. DiffRate\cite{yu2023diffrate} addresses the challenge of selecting an optimal merge ratio by rendering the rate itself differentiable. It utilizes a learnable budget controller to optimize this rate for each input, facilitating instance-adaptive efficiency through standard gradient descent but increasing complexity. ToFu\cite{song2024tofu} diverges from ToMe's training-free methodology by proposing a learnable fusion module that is co-trained with the models to generate new, more expressive tokens. Hybrid approaches such as Pumer\cite{fu2024pumer} and LTPM\cite{li2024ltpm} integrate token pruning and merging within a unified framework. Pumer introduces a learnable router to dynamically determine the number of tokens to prune and merge on a per-instance basis, whereas LTPM employs learnable parameters to decide whether a token should be pruned or which tokens should be merged. Extending beyond pure vision models, recent works like PruMerge \cite{shang2025prumerge} and VisionZip \cite{yang2025visionzip} propose highly effective hybrid strategies for MLLMs. They typically first prune inattentive visual tokens based on attention scores and subsequently merge the remaining redundant tokens using spatial or semantic similarity matching.

\subsubsection{Clustering-Based Reduction}
Clustering approaches use offline algorithms to group similar tokens. TCFormer\cite{tcformer} employs KNN-enhanced Density Peaks Clustering to group tokens and merge redundant ones through averaging for human activity tasks like pose estimation. ClusTR\cite{clustr} uses hierarchical token merging with cosine similarity across Transformer layers for vision tasks, but its fixed ratios limit flexibility and may hinder small object detection. For handling severe spatiotemporal redundancy in long videos, Chat-UniVi \cite{jin2024chatunivi} successfully adapts Density Peak Clustering (DPC-KNN) to group semantically similar frame representations, unifying image and video token compression within a single framework. While these methods preserve global context, they face three drawbacks: 1) Iterative clustering algorithms with O(nk) complexity offset computational gains, 2) Discrete cluster assignments prevent gradient flow, and 3) Fixed cluster counts lack input adaptability.

\subsubsection{Learnable Token Reduction}
End-to-end trainable methods optimize reduction policies through differentiable architectures. ATS\cite{fayyaz2022ats} implements token merging via weighted averaging with gating mechanisms. Dynamic Token Morphing\cite{wang2023morphing} uses cross-attention between original and learnable proxy tokens for information absorption. Gumbel Token Selector\cite{kim2023gumbel} employs Gumbel-Softmax to sample token subsets through residual connections. These approaches show promise but increase model complexity (15-30\% more parameters) and risk overfitting on small datasets. In the context of long-context multimodal processing, query-based token distillation has emerged as a powerful paradigm. For instance, LLaMA-VID \cite{li2024llamavid} compresses each video frame into a minimal set of learnable tokens guided by text prompts, achieving extreme token reduction while preserving crucial contextual semantics. 

\section{Methodology}
\label{sec:method}

\subsection{Merging Process}
\paragraph{Token Partitioning}
Let the input sequence from a given layer be represented by the matrix $X \in \mathbb{R}^{L \times d}$, where $L$ is the number of tokens and $d$ is the feature dimension. We first partition this sequence into two disjoint sets: a set of $M$ destination tokens, denoted by $\mathbf X_{\text{dst}} \in \mathbb{R}^{M \times d}$, and a set of $N$ source tokens, $\mathbf X_{\text{src}} \in \mathbb{R}^{N \times d}$, where $L=M+N$. 
\begin{align}
\mathbf X_{\text{dst}} &= \{\mathbf{x}_i : i \in \mathcal{I}_{\text{dst}}\} \notag \\
\mathbf X_{\text{src}} &= \{\mathbf{x}_j : j \in \mathcal{I}_{\text{src}}\}
\end{align}
where $\mathcal{I}_{dst}$ and $\mathcal{I}_{src}$ represent the index sets for destination and source tokens, respectively, such that $\mathcal{I}_{dst} \cap \mathcal{I}_{src} = \emptyset$ and $\mathcal{I}_{dst} \cup \mathcal{I}_{src} = \mathcal{I}$ covers all token indices, excluding any special tokens (e.g., class tokens).
The specific strategy for partitioning into $\mathcal{I}_{\text{dst}}$ and $\mathcal{I}_{\text{src}}$ can vary (e.g., fixed interleaved patterns or random selection).

\paragraph{Similarity-Based Fusion Matrix}
We begin by computing the cosine similarity between each destination token and every source token. This yields a similarity matrix $S \in \mathbb{R}^{M \times N}$, where each element $S_{ij}$ is defined as:
\begin{equation}
S_{ij} = \frac{\mathbf{x}_i^{\text{dst}} \cdot \mathbf{x}_j^{\text{src}}}{\|\mathbf{x}_i^{\text{dst}}\| \cdot \|\mathbf{x}_j^{\text{src}}\|}
\label{eq:sim}
\end{equation}
To isolate the most significant relationships, we apply a rectified linear unit (ReLU) activation with a shifting threshold $\tau$. This step filters out weak connections, producing a sparse similarity matrix $\tilde{S} \in \mathbb{R}^{M \times N}$:
\begin{equation}
\tilde{S}_{ij} = \text{ReLU}(S_{ij} - \tau)
\end{equation}

\paragraph{Adaptive Weight Refining.} From the sparse similarity matrix $\tilde{S}$, we first compute an initial weight matrix $W \in \mathbb{R}^{M \times N}$ by normalizing its columns. This ensures the initial influence of each source token is properly distributed among its similar destination tokens.
\begin{equation}
W_{ij} = \frac{\tilde{S}_{ij}}{\sum_{i=1}^{M} \tilde{S}_{ij} + \epsilon}
\end{equation}
where $\epsilon$ is a small constant for numerical stability.

To further refine these weights, we introduce a dynamic, column-specific thresholding mechanism. For each source token $j$, we define a threshold $\zeta_j$ as the average of its non-zero weights in $W$:

\begin{equation}
\zeta_j = \frac{\sum_{i=1}^{M} W_{ij}}{C_{j} + \epsilon}
\end{equation}
where $C_{j}$ is the count of non-zero entries along the destination dimension and can be computed as

\begin{equation}
C_{j}= \sum_{i=1}^{M} \frac{W_{ij}}{W_{ij} + \epsilon}
\end{equation}

The threshold $\zeta_j$ is to prune connections that are weak relative to a source token's other connections. We apply this threshold to obtain a pruned weight matrix $\tilde W$:

\begin{equation}
\tilde{W}_{ij} = \text{ReLU}(W_{ij} - \zeta_j)
\end{equation}
Finally, the pruned matrix $\tilde{W}$ is re-normalized column-wise to produce the final fusion weights $W^{\text{F}} \in \mathbb{R}^{M \times N}$:
\begin{equation}
W^{\text{F}}_{ij} = \frac{\tilde W_{ij}}{\sum_{i=1}^{M} \tilde W_{ij} + \epsilon}
\end{equation}

\paragraph{Token Aggregation and Preservation}
The destination tokens are updated by aggregating the features from source tokens, guided by the final fusion weights. The fused destination tokens, $X''_{\text{dst}} \in \mathbb{R}^{M \times d}$, are computed as:

\begin{align}
\mathbf{X}'_{\text{dst}} &= \mathbf{X}_{\text{dst}} + \mathbf{W}^{\text{F}} \mathbf{X}_{\text{src}} \notag \\
\mathbf{x}''_{\text{dst}, i} &= \frac{\mathbf{x}'_{\text{dst}, i}}{1 + \sum_{j=1}^{N} W^{\text{F}}_{ij}} \label{eq:fusion}
\end{align}

A key component of our methodology is the preservation of unique source tokens $\mathbf{X}_{\text{pres}}$. A source token $x_j^\text{src}$ is preserved if it exhibits no similarity to any destination token, which means the sum of its similarities to all $M$ destination tokens is zero: $m_j =\mathbb{I}( \sum_{i=1}^{M} {W}_{ij}^{F} =0)$, where $\mathbb{I}(\cdot)$ is the indicator function. So $\mathbf{X}_{\text{pres}}=\{\mathbf{x}_{j}^{src} \mid m_j = 1\}$. 

The final reduced sequence $\mathbf X'$ is formed by concatenating any special tokens $\mathbf X_{\text{spec}}$, the fused destination tokens $\mathbf X''_{\text{dst}}$, and the set of preserved source tokens $\mathbf X_{\text{pres}}$.
\begin{equation}
\mathbf{X}' = \text{concat}(\mathbf{X}_{\text{spec}}, \mathbf{X}''_{\text{dst}}, \mathbf{X}_{\text{pres}})
\end{equation}
If $r$ source tokens satisfy the preservation condition and there are $l_{\text{spec}}$ special tokens, the resulting sequence will have a reduced length of $l_{\text{spec}}+M+r$.

\paragraph{Batch Processing Implementation}  For efficient implementation on batched data, the preservation decision must be consistent across all samples in a batch. Given the fusion matrix for a batch be $\mathbf{W}^F \in \mathbb{R}^{B \times M \times N}$. A per-sample preservation mask $m^{(b)} \in \{0,1\}^{N}$ is computed for each sample $b$, where $m^{(b)}_j = \mathbb{I}(\sum_{i=1}^{M} W^F_{bij} = 0)$. To ensure batch consistency, a source token $j$ is preserved if it is marked for preservation in \textit{any} sample, yielding a final batch-wide mask $m^{\text{final}}_j = \bigvee_{b=1}^{B} m^{(b)}_j$. So the preserved source tokens $\mathbf{X}_{\text{pres}}=\{\mathbf{x}_{j}^{src} \mid m^{\text{final}}_j=1\}$. Subsequently, to prevent preserved tokens from fusing, the corrected fusion matrix $\mathbf{\tilde{W}}^{F}$ is obtained by $\mathbf{\tilde{W}}^{F}_{b,i,j} = \mathbf{W}^{F}_{b,i,j} \cdot (1 - m^{\text{final}}_j)$, zeroing out columns corresponding to preserved tokens and keeping others unchanged.

\subsection{Computational Efficiency}
Our token merging method aims to reduce the computational cost of self-attention by effectively shortening the sequence length. The overhead introduced by the merging process itself is analyzed as follows:
\begin{itemize}
\item \textbf{Similarity Matrix Calculation ($S$):} Computing cosine similarity~\ref{eq:sim} between $M$ destination tokens and $N$ source tokens involves a matrix multiplication of shape $(M \times d)$ with $(d \times N)$, resulting in $O(M N d)$ operations.
\item \textbf{Adaptive Weight Pruning ($W$, $\zeta_j$, $\tilde W$, $W^{\text{F}}$):} These involve several matrix operations, primarily column-wise summations and element-wise operations. These steps are dominated by the $O(M N)$ complexity of iterating through the similarity matrix.
\item \textbf{Token Aggregation ($\mathbf{X}'_\text{dst}$):} The aggregation~\ref{eq:fusion} involves a matrix multiplication of shape $(M \times N)$ with $(N \times d)$, resulting in $O(M N d)$ operations.
\item \textbf{Token Preservation:} Identifying preserved tokens involves column-wise summation on $\tilde{S}$, which is $O(M N)$.
\end{itemize}
 Given that $M$ and $N$ are fractions of the original sequence length $L$ (i.e., $M \approx \alpha L$, $N \approx (1-\alpha)L$), this overhead scales approximately as $O(\alpha(1-\alpha)L^2 d)$, similar to self-attention. When merging is applied, the subsequent attention computation is reduced to $O(L'^2 d)$. Assuming $L' \approx \beta L$ with $\alpha \leq \beta \leq 1$, this becomes $O(\beta^2 L^2 d)$. Therefore, the total cost is $O\left( \left( \alpha(1-\alpha) + \beta^2 \right) L^2 d \right)$. For more efficient than standard self-attention, it requires $\alpha(1-\alpha) + \beta^2 < 1$, which simplifies to the condition $\beta < \sqrt{\alpha^2 - \alpha + 1}$. To assess the strictness of this condition, we consider the case where $\alpha$ and $\beta$ are uniformly distributed over $(0,1)$ with $\alpha \leq \beta$. The probability that $\beta < \sqrt{\alpha^2 - \alpha + 1}$ is given by: The area of the region $A = { (\alpha, \beta) \mid 0 < \alpha \leq \beta \leq 1 }$ is $\frac{1}{2}$; the area of the region $B = { (\alpha, \beta) \in A \mid \beta < \sqrt{\alpha^2 - \alpha + 1} }$ is $\int_{0}^{1} \left( \sqrt{\alpha^2 - \alpha + 1} - \alpha \right) d\alpha = \frac{3}{8} \ln 3$. Therefore, the probability is $P = B/A= \frac{3}{4} \ln 3 \approx 0.824,$, indicating that the condition holds in approximately 82.4\% of cases. This means that for most parameter choices, it achieves computational efficiency. Moreover, even if the condition is not strictly met in the current block, the reduced sequence length $L'$ propagates to subsequent blocks, ensuring that all following attention computations benefit from the shorter sequence, leading to overall computational savings across the network.

\subsection{Restoration Process}
The restoration process aims to reconstruct the original sequence $X \in \mathbb{R}^{L \times d}$ from the reduced sequence $X'$. This process utilizes the stored fusion matrix $W^{\text{F}} \in \mathbb{R}^{M \times N}$ from the merging step. 
\\
\textbf{Split the Merged Sequence:} The input to the unmerging process is $X'$, which consists of special tokens $X_{\text{spec}}$, $X'_{\text{dst}}$, and $X_{\text{pres}}$. We first separate these components. Let $X'_{\text{dst}} \in \mathbb{R}^{M \times d}$ be the fused destination tokens and $X_{\text{pres}} \in \mathbb{R}^{r \times d}$ be the adaptively preserved source tokens.
\\
\textbf{Reconstruct Source Tokens ($X^{\text{rec}}_{\text{src}}$):} For the source tokens that were adaptively preserved, they are directly taken from $X_{\text{pres}}$.
For source tokens that were not preserved (i.e., those that were merged into destination tokens), the ideal reconstruction would require inverting the fusion process\ref{eq:fusion}. For a non-square matrix $W^{\text{F}}$, a principled choice is to get its Moore–Penrose pseudo-inverse, which provides the minimum-norm least-squares solution but is expensive. Intuitively, the unmerging process can be understood as: \emph{a source token retrieves a similar proportion of information it originally contributed to the merged destination tokens.} Therefore, we simply aggregate the corresponding fractions of every fused destination token to which the source token originally contributed, utilizing the weights stored in matrix $W^{\text{F}}$.  Each reconstructed source token $\mathbf{x}_{src,j}^{\text{rec}}$ is computed as:
\begin{equation}
\mathbf{x}_{src, j}^{\text{rec}} = \sum_{i=1}^{M} \left( \frac{W^{\text{F}}_{ji}}{R_i} \right) \mathbf{X}'_{\text{dst}, i}
\end{equation}
where $R_i = 1 + \sum_j W^{\text{F}}_{ij}$. This term $\frac{W^{\text{F}}_{ji}}{R_i}$ represents the proportion of the $j$-th source token's contribution to the $i$-th merged destination token, scaled by the total weight of the $i$-th merged token during the merging process. This effectively distributes the information from fused destination tokens back to their original source token positions.
\\
\textbf{Reconstruct Destination Tokens ($X^{\text{rec}}_{\text{dst}}$):} The original destination tokens $\mathbf{x}_{dst,i}^{\text{rec}}$ are reconstructed from the merged destination tokens $\mathbf{X}'_{\text{dst}, i}$ is effectively performed as:
\begin{equation}
\mathbf{x}_{dst,i}^{\text{rec}} = \frac{\mathbf{X}'_{\text{dst}, i}}{R_i}
\end{equation}
This step aims to retrieve the original destination tokens by de-scaling the merged output.
\\
\textbf{Assemble the Original Sequence:} Finally, the preserved tokens ${X}_{\text{pres}}$, the reconstructed destination tokens $X^{\text{rec}}_{\text{dst}}$, the reconstructed source tokens $X^{\text{rec}}_{\text{src}}$, and any special tokens $X_{\text{spec}}$ are reassembled into their original sequence order to recover $X^{\text{rec}}$.
\begin{equation}
\mathbf{X}^{\text{rec}} = \text{concat}(\mathbf{X}_{\text{spec}}, \mathbf{X}^{\text{rec}}_{\text{src}},\mathbf{X}^{\text{rec}}_{\text{dst}},\mathbf{X}_{\text{pres}})
\end{equation}

\subsection{Integration with Transformers}

Our proposed MaMe module, denoted as $\text{MaMe}(\cdot)$, and its inverse, $\text{MaRe}(\cdot)$, are seamlessly integrated into the transformer-based architecture. Let $x_{l-1}$ be the token sequence output by block $l-1$, and $\text{MSA}$, $\text{MLP}$, and $\text{LN}$ denote Multi-head Self-Attention, Multi-Layer Perceptron, and Layer Normalization, respectively.

\paragraph{Integration for Visual Perception}

For visual representation tasks, the primary objective is to maximize computational efficiency without incurring significant accuracy degradation. The operations within a modified Transformer block $l$ for classification are formalized as follows:
\begin{align}
x'_{l} &= \text{MSA}(\text{LN}(x_{l-1})) + x_{l-1} \\
x''_{l} &= \text{MaMe}(x'_{l}) \\
x_{l} &= \text{MLP}(\text{LN}(x''_{l})) + x''_{l}
\label{eq:gen}
\end{align}

\paragraph{Integration for Image Synthesis}

For image generation tasks where maintaining the original token count or spatial resolution throughout the network is critical. The operations within a modified Transformer block $l$ for generation are outlined as follows:
\begin{align}
x'_{l} &= \text{MaRe}(\text{MSA}(\text{MaMe}(\text{LN}(x_{l-1})))) + x_{l-1} \\
x_{l} &= \text{MLP}(\text{LN}(x'_{l})) + x'_{l}
\end{align}

\begin{algorithm}[t!]
\caption{Matrix-Based Token Merging}
\label{alg:merging}
\small
\begin{algorithmic}[1]
\Require Input tokens $X \in \mathbb{R}^{L \times d}$, similarity threshold $\tau$
\Ensure Reduced and fused token sequence $X' \in \mathbb{R}^{L' \times d}$
\State \textbf{Note:} $X_{\text{spec}}$ refers to any special tokens (e.g., class tokens) that are preserved without modification.

\Function{MaMe}{$X, \tau$}
    \State Partition $X$ into destination tokens $X_{\text{dst}} \in \mathbb{R}^{M \times d}$ and source tokens $X_{\text{src}} \in \mathbb{R}^{N \times d}$
    \State \textbf{Step 1. Compute Similarity Matrix}
    \State Initialize similarity matrix $S \in \mathbb{R}^{M \times N}$
    \For{$i = 0 \to M-1$}
        \For{$j = 0 \to N-1$}
            \State $S_{ij} \leftarrow \frac{\mathbf{x}_{i}^{\text{dst}} \cdot \mathbf{x}_{j}^{\text{src}}}{\|\mathbf{x}_{i}^{\text{dst}}\| \cdot \|\mathbf{x}_{j}^{\text{src}}\|}$
        \EndFor
    \EndFor
    \State $\tilde{S} \leftarrow \text{ReLU}(S - \tau)$

    \State \textbf{Step 2. Compute Fusion Weights}
    
    \State Initialize $W, \tilde W, W^{\text{F}} \in \mathbb{R}^{M \times N}$ (all zeros initially)

    \For{$j = 0 \to N-1$}
            \For{$i = 0 \to M-1$}
                \State $W_{ij} \leftarrow \tilde{S}_{ij} / (\sum_{i=0}^{M-1} \tilde{S}_{ij} + \epsilon)$
            \EndFor

            \State $C_{j} \leftarrow \sum_{i=0}^{M-1} \mathbb{I}(W_{ij} > 0)$
            \State $\zeta_j \leftarrow (\sum_{i=0}^{M-1} W_{ij}) / (C_{j} + \epsilon)$
            
            \For{$i = 0 \to M-1$}
                \State $\tilde W_{ij} \leftarrow \text{ReLU}(W_{ij} - \zeta_j)$
            \EndFor

            \State $\text{Mask} \leftarrow \sum_{i=0}^{M-1} \tilde W_{ij}$

            \For{$i = 0 \to M-1$}
                    \State $W^{\text{F}}_{ij} \leftarrow W_{ij} / (\text{Mask}_{j} + \epsilon)$
            \EndFor
            
    \EndFor

    \State \textbf{Step 3. Compute Preserved and Fusion Tokens}
    \State Initialize $X_{\text{pres}} \leftarrow \emptyset$
    \State Initialize fused destination tokens $X'_{\text{dst}} \in \mathbb{R}^{M \times d}$
    \For{$j = 0 \to N-1$}
        \If{$\text{Mask}_{j} = 0$}
        \State Add $\mathbf{x}_j^{\text{src}}$ to $X_{\text{pres}}$
        \EndIf
    \EndFor
    \For{$i = 0 \to M-1$}
        \State $\mathbf{x}'_{\text{dst}, i} \leftarrow \mathbf{X}_i^{\text{dst}} + \sum_{j=0}^{N-1} 
        W^{\text{F}}_{ij} \mathbf{x}_j^{\text{src}}$
        \State $\mathbf{x}''_{\text{dst}, i} = \mathbf{x}'_{\text{dst}, i}/(1 + \sum_{j=1}^{N} W^{\text{F}}_{ij})$
    \EndFor

    \State \textbf{Step 4. Assemble the Final Tokens}
    \State $X' \leftarrow \text{concat}(X_{\text{spec}}, X''_{\text{dst}}, X_{\text{pres}})$
    \Comment{Concatenate the special, fused destination, and preserved tokens}
    
    \State \textbf{return} $X'$
\EndFunction
\end{algorithmic}
\end{algorithm}

\begin{algorithm}[!ht]
\caption{Matrix-Based Token Restoration}
\label{alg:restore}
\small
\begin{algorithmic}[1]
\Require Reduced sequence $X' \in \mathbb{R}^{L' \times d}$,
         Fusion matrix $W^{\text{F}} \in \mathbb{R}^{M \times N}$,  \text{mask}$\in \{0,1\}^{N}$, Length $l_{\text{spec}}$
\Ensure Reconstructed original sequence $X_{\text{rec}} \in \mathbb{R}^{L \times d}$

\Function{Restoration}{$X', W^{\text{F}}, \text{mask}, l_{\text{spec}}$}
    \State \textbf{Step 1. Split the Merged Tokens}
    \State Extract $X_{\text{spec}} \leftarrow X'[0 \dots l_{\text{spec}}-1, :]$
    \State Extract $X'_{\text{dst}} \leftarrow X'[l_{\text{spec}} \dots l_{\text{spec}}+M-1, :]$
    \State Extract $X_{\text{pres}} \leftarrow X'[l_{\text{spec}}+M \dots L'-1, :]$

    \State \textbf{Step 2. Reconstruct Destination Tokens ($X_{\text{dst}}$)}
    \State Initialize $X^{\text{rec}}_{\text{dst}} \in \mathbb{R}^{M \times d}$ (all zeros initially)

    \For{$i = 0 \to M-1$}
        \State $ R_{i} \leftarrow \sum_{j=1}^N W^{\text{F}}_{ij} + 1 $
        \State $\mathbf{x}_{dst,i}^{\text{rec}} \leftarrow \mathbf{x}'_{\text{dst}, i} / R_{i}$
    \EndFor

    \State \textbf{Step 3. Reconstruct Source Tokens ($X_{\text{src}}$)}
    \State Initialize $X^{\text{rec}}_{\text{src}} \in \mathbb{R}^{N \times d}$ (all zeros initially)

    \For{$j = 0 \to N-1$}
        \If{$\text{mask}_j = 0$}
            \State $\mathbf{x}_{src, j}^{\text{rec}} \leftarrow \sum_{i=0}^{M-1} {W^{\text{F}}_{ji}} \mathbf{x}^{\text{rec}}_{\text{dst}, i}$
           
        \Else
            \State $\mathbf{x}_{src, j}^{\text{rec}} \leftarrow \text{corresponding token from } X_{\text{pres}}$
           
        \EndIf

    \EndFor

    \State \textbf{4. Assemble the Original Sequence}
    \State Initialize $X_{\text{rec}} \in \mathbb{R}^{L \times d}$ (all zeros initially) 
    \State $X^{\text{rec}} \leftarrow \text{concat}(X_{\text{pres}}, X_{\text{spec}}, X^{\text{rec}}_{\text{dst}}, X^{\text{rec}}_{\text{src}})$
    \Comment{Place  $X_{\text{pres}}, X_{\text{spec}}, X^{\text{rec}}_{\text{dst}}, X^{\text{rec}}_{\text{src}}$ back into their original positions}
    \State \textbf{return} $X_{\text{rec}}$
\EndFunction
\end{algorithmic}
\end{algorithm}

\section{Experiments}

\subsection{Image Classification}
We conduct comprehensive experiments on the ImageNet-1K dataset, a visual recognition benchmark containing 1.28 million training images and 50,000 validation images across 1,000 object categories. We report Top-1 accuracy for performance comparison and throughput for efficiency measurement.

\subsubsection{Implementation Details}

\begin{figure*}[!t]
\centering

\includegraphics[width=\textwidth]{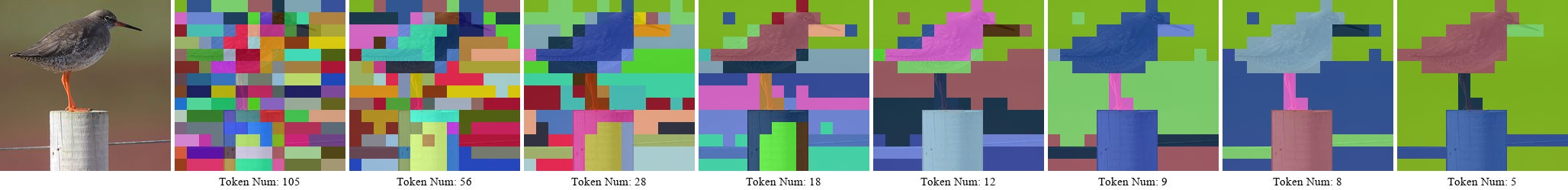}%
\\

\includegraphics[width=\textwidth]{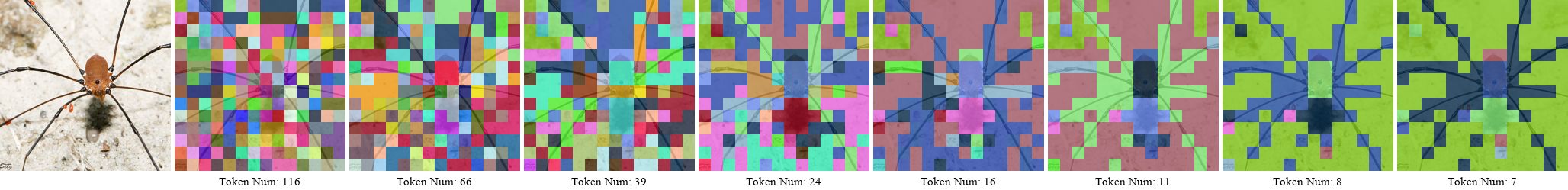}%
\\
\includegraphics[width=\textwidth]{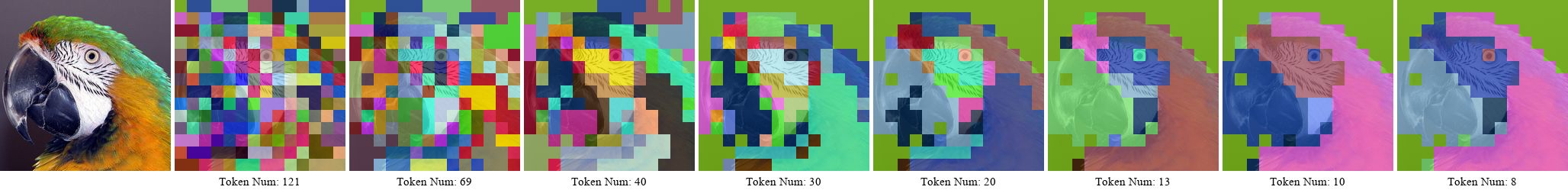}%
\\
\includegraphics[width=\textwidth]{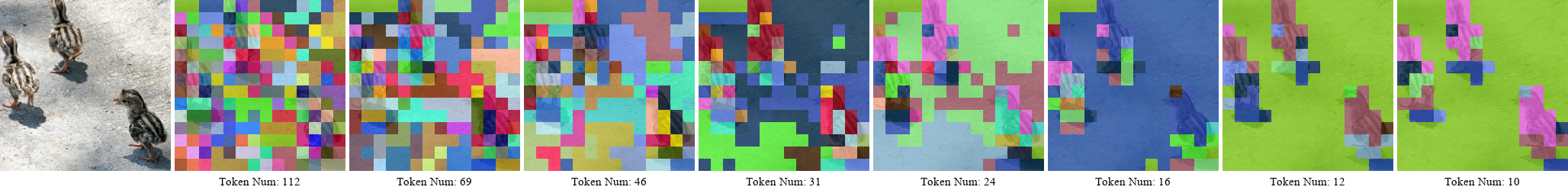}%

\vspace{-5pt}
\caption{The visualization illustrates the progression of token count reduction in the first 8 blocks of the AugReg ViT-B/16 with MaMe. Each color square represents a distinct type of token. More results are in the supplementary material.}
\label{fig:visprogress}
\end{figure*}

\paragraph{Training-Free}
The evaluation employs two representative architectures: DeiT\cite{touvron2021training} and MAE\cite{he2022masked}, utilizing their pre-trained weights without any fine-tuning. For these off-the-shelf experiments, we apply MaMe to the first 8 layers of each model, where we empirically set the similarity threshold to 0.8. All other things remain identical to \cite{yu2023diffrate}.

\paragraph{Training-From-Scratch}
For end-to-end training experiments, we follow the training recipes~\cite{liu2021swin} while incorporating our compression strategy. In standard ViT architectures, we set the similarity threshold 0.5 and apply merging at layers 3, 6, and 9, implementing a consistent 2:1 token reduction ratio (reducing token count, except for the special class token, to half of original) at each compression point. . All other training settings including optimization (AdamW), and learning rate schedule (cosine decay with 20-epoch warmup) remain identical to~\cite{liu2021swin}.

\paragraph{Fine-tuning}
To assess the adaptability of MaMe on pre-trained models, we investigate two compression configurations under a fine-tuning protocol. The first setting applies MaMe at layers 3, 6, and 9, mirroring the from-scratch setup. The second setting restricts compression exclusively to the final transformer block. Both configurations are fine-tuned for 1 epoch to adapt the compressed architecture to the pre-trained weights. All other training settings, including optimization (AdamW) and learning rate schedules, remain identical to ~\cite{liu2021swin}.

\subsubsection{Results}

\begin{table}[ht!]
    \centering
    \caption{The Results of Token compression on the off-the-shelf models. Throughput is measured on an A100 GPU with a batchsize of 1024 using FP16 precision.}
    \label{tab:free}
    \small
    \begin{tabular}{ccccc}
    \toprule
    \multirow{2}{*}{Model}             & \multirow{2}{*}{Method}       & FLOPs          & Throughput     & Top-1 Acc  \\ 
                                              & & (G)            & (img/s)        & (\%)       \\   \midrule
    \multicolumn{5}{c}{\textbf{Training Free on ImageNet-1K (224$\times$224)}}                                                                                      \\   \midrule
     \multirow{5}{*}{\shortstack{ViT-S (DeiT)}}  
              &  Baseline  & 4.6  & 5039  & 79.82   \\
              &  EViT & 2.3  & 8950  &  73.83 \\
              &  ToMe & 2.3  & 8874  &  77.99 \\
              &  DiffRate &  2.3 & 8875  &  78.75\\ 
               & \cellcolor{Gray}  MaMe & \cellcolor{Gray} 2.3  &\cellcolor{Gray}  9015&\cellcolor{Gray}  78.61 \\
    \midrule 
    \multirow{5}{*}{ViT-B (DeiT)}  
              &  Baseline  & 17.6  & 2130  & 81.83   \\
              &  EViT & 8.7  & 4230  & 74.61    \\
              &  ToMe &  8.8 & 4023  & 77.84  \\
              &  DiffRate &  8.7 & 4124  &  78.98 \\
       & \cellcolor{Gray}  MaMe  & \cellcolor{Gray} 8.7  &\cellcolor{Gray} 4117  & \cellcolor{Gray}79.80  \\
    \midrule 
    \multirow{5}{*}{ViT-B (MAE)} 
              &  Baseline  & 17.6  & 2130  & 83.72   \\
              &  EViT & 8.7  & 4230  & 75.15    \\
              &  ToMe  & 8.8  & 4023 & 78.86    \\ 
              &  DiffRate &  8.7 & 4150  &  79.96 \\
    &\cellcolor{Gray}   MaMe  &\cellcolor{Gray} 8.7  &\cellcolor{Gray} 5418  &\cellcolor{Gray}  79.83 \\
              
    \midrule 
    \multirow{5}{*}{ViT-L (MAE)} 
              &  Baseline  & 61.6  & 758  & 85.95   \\
              &  EViT & 29.7  & 1672  & 81.52    \\
              &  ToMe  & 31.0  & 1550 & 84.24    \\
              &  DiffRate &  31.0 & 1580  &  84.65 \\
    &\cellcolor{Gray}   MaMe  &\cellcolor{Gray} 31.0  &\cellcolor{Gray} 2764  &\cellcolor{Gray} 84.81  \\
    \midrule 
    \multirow{5}{*}{ViT-H (MAE)} 
              &  Baseline  & 167.4  & 299  & 86.88   \\
              &  ToMe  &  92.9 & 500 &  86.01   \\ 
              &  EViT & 99.1  & 512  & 85.54    \\
              &  DiffRate &  93.2 & 504  &  86.40 \\
     &\cellcolor{Gray}  MaMe &\cellcolor{Gray} 93.2  &\cellcolor{Gray}  908 &\cellcolor{Gray}  85.51  \\
    \hline 
    \end{tabular}

\end{table}

\paragraph{Training-Free}  
Table~\ref{tab:free} evaluates several token compression methods on ViT models. For ViT-S (DeiT), MaMe achieves 9015 img/s, 79\% higher than baseline while maintaining 78.61\% accuracy (1.2 points below original), surpassing EViT (8950 img/s, 73.83\% accuracy) and ToMe (8874 img/s, 77.99\% accuracy). For ViT-B (DeiT), MaMe delivers 4117 img/s (93\% faster) with 79.80\% accuracy. EViT shows higher throughput (4230 img/s) but lower accuracy (74.61\%), while DiffRate\cite{yu2023diffrate} has similar speed but 78.98\% accuracy. 

Notably, comparing ViT-B (DeiT) and ViT-B (MAE), which share identical architecture, reveals significant differences in MaMe's performance. On ViT-B (DeiT), MaMe achieves 4117 imgs/s with 79.80\% accuracy, while on MAE, throughput increases to 5418 imgs/s, representing a 31.6\% throughput improvement, while maintaining 79.83\% accuracy. This highlights MaMe's ability to leverage MAE's self-supervised representations more effectively than supervised ones. EViT and ToMe show no throughput change between ViT-B models. MaMe's advantage grows with model size: for ViT-L (MAE), it achieves 2764 imgs/s (EViT's 1.63x) while maintaining the highest accuracy (84.81\%). On ViT-H (MAE), MaMe delivers 908 imgs/s (almost EViT's 2x), with only a marginal accuracy decrease compared to DiffRate.

\paragraph{Traing From Scratch} 
Table~\ref{tab:nofree} shows results for training-from-scratch ViT models with MaMe compression at layers 3/6/9 (marked $^\dagger$). MaMe consistently enhances throughput across all model scales while maintaining competitive accuracy. ViT-T$^\dagger$ achieves 4462 img/s, nearly doubling the baseline throughput (2291 img/s) with only a 1.3 percentage point accuracy drop (70.9\% vs. 72.2\%). For larger models, ViT-B$^\dagger$ reaches 813 img/s (93\% speedup) at a 5.8-point accuracy cost (76.0\% vs. 81.8\%). However, compared to fine-tuning, it is not recommended to train from scratch.

\paragraph{Fine-tuning}
We evaluate two fine-tuning strategies: MaMe at layers 3/6/9 ($^\ddagger$) (mirroring to training-from-scratch) versus compression at the final transformer block ($^\star$). Two consistent trends emerge in Table~\ref{tab:nofree}. First, fine-tuning consistently outperforms from-scratch training: ViT-S$^\ddagger$ (78.2\%) exceeds ViT-S$^\dagger$ (77.0\%) by +1.2 pp, and ViT-B$^\ddagger$ (77.7\%) surpasses ViT-B$^\dagger$ (76.0\%) by +1.7 pp, indicating better adaptation to pre-trained representations.

Second, last-layer compression ($^\star$) improves accuracy beyond uncompressed baselines: ViT-T$^\star$ achieves 72.4\% (+0.2 pp), ViT-S$^\star$ reaches 80.2\% (+0.4 pp), and ViT-B$^\star$ attains 82.8\% (+1.0 pp). This suggests selective compression acts as an implicit regularizer, pruning redundant tokens while preserving discriminative features.Overall, fine-tuning or last-layer-fine-tuning is recommended.

\begin{table}[ht!]
  \centering
    \caption{The Results of Vision Architectures with MaMe Compression. The mark $^\dagger$ denotes MaMe compression applied to three blocks (layers 3/6/9) with training from scratch; $^\ddagger$ indicates MaMe applied to three blocks (layers 3/6/9) followed by fine-tuning; $^\star$ denotes MaMe applied only to the last transformer block followed by fine-tuning. Throughput is measured on an A100 GPU with a batch size of 64 using FP32 precision.}
  \label{tab:nofree}
  \footnotesize
  \resizebox{0.97\columnwidth}{!}{
  \begin{tabular}{ccccc} 
  \toprule
  \multirow{2}{*}{Model}             & Param         & FLOPs          & Throughput     & Top-1 Acc  \\ 
                                    & (M)           & (G)            & (img/s)        & (\%)       \\   \midrule
  
  \multicolumn{5}{c}{\textbf{ImageNet-1K Results (224$\times$224)}} \\   \midrule

   ViT-T & 5.72 & 1.3 & 2291 & 72.2 \\ 
   ViT-T$^\dagger$ & 5.72 & 0.6 & 4462 & 70.9 \\ 
   ViT-T$^\ddagger$ & 5.72 & 1.0 & 4462 & 71.2 \\ 
   ViT-T$^\star$ & 5.72 & 1.2 & 2468 & 72.4 \\ 
   
   \midrule
   
   ViT-S & 22.0 & 4.6 & 1157 & 79.8 \\
   ViT-S$^\dagger$ & 22.0 & 2.1 & 2257 & 77.0 \\
   ViT-S$^\ddagger$ & 22.0 & 2.1 & 2257 & 78.2 \\
   ViT-S$^\star$ & 22.0 & 4.1 & 1250 & 80.2 \\
   
   \midrule
   
   ViT-B & 86.4 & 17.6 & 422 & 81.8 \\
   ViT-B$^\dagger$ & 86.4 & 8.4 & 813 & 76.0 \\
   ViT-B$^\ddagger$ & 86.4 & 8.4 & 813 & 77.7 \\
   ViT-B$^\star$ & 86.4 & 16.0 & 460 & 82.8 \\

   \bottomrule
  \end{tabular}
  } 
\end{table}

\subsection{Multimodal Large Language Models}

\paragraph{Zero-shot Image Classification}
We conducted zero-shot image classification on ImageNet-1K validation set to evaluate token merging strategies across CLIP\cite{radford2021learningtransferablevisualmodels}, SigLIP\cite{zhai2023sigmoid}, and SigLIP2\cite{tschannen2025siglip2}. For CLIP, MaMe ($\tau=0.8$) increased throughput by 25\% (64.01 img/s) with 0.39\% accuracy drop, while ToMe (r=12) gave 3\% throughput gain with 4.34\% accuracy loss. For SigLIP, MaMe ($\tau=0.9$) improved throughput by 25\% (58.10 img/s) with 1.11\% accuracy reduction, while ToMe (r=32) achieved 19\% speedup with 1.28\% accuracy loss. For SigLIP2, MaMe ($\tau=0.95$) increased throughput by 28\% (56.15 img/s) with 0.35\% accuracy drop, while ToMe (r=32) gave 16\% throughput gain with 1.91\% accuracy loss. SigLIP2's ability to merge tokens at $\tau=0.95$ indicates its confident semantic representations. MaMe demonstrates better balance between throughput and accuracy versus baseline and ToMe.

\begin{table}[htbp!]
\centering
\caption{Zero-shot image classification results. Inference throughput measured on a 3090 GPU with FP32 precision.}
\label{tab:zero}
\small
\renewcommand\arraystretch{1.5}
\resizebox{\columnwidth}{!}{
 \begin{tabular}{ccccc}
        \toprule
        \multirow{2}{*}{Model}      & \multirow{2}{*}{Method} & Input Size & Throughput & Top-1 Acc  \\ 
                                    &                         & (px)       & (img/s)    & (\%)       \\ \midrule
        \multicolumn{5}{c}{\textbf{Zero-Shot Classification on ImageNet-1K}} \\ \midrule
        \multirow{5}{*}{CLIP (ViT-L/14)} 
                                    & Baseline                & 224        & 51.22        & 70.34     \\
                                    & ToMe(r=8)                    & 224        & 51.33       & 68.98       \\ 
                                    & ToMe(r=12)                   & 224        & 52.86       & 66.00       \\ 
                                    & \cellcolor{Gray} MaMe($\tau=0.7$) & \cellcolor{Gray} 224 & \cellcolor{Gray} 69.09 & \cellcolor{Gray} 67.60  \\ 
                                    & \cellcolor{Gray} MaMe($\tau=0.8$) & \cellcolor{Gray} 224 & \cellcolor{Gray} 64.01 & \cellcolor{Gray} 69.95  \\ \midrule
        \multirow{5}{*}{SigLIP (ViT-B/16)} 
                                    & Baseline                & 512        & 46.28        & 75.61       \\
                                    & ToMe(r=32)              & 512        & 55.10       & 74.33       \\ 
                                    & ToMe(r=64)              & 512        & 71.94       & 70.66       \\ 
                                    & \cellcolor{Gray} MaMe($\tau=0.8$) & \cellcolor{Gray} 512 & \cellcolor{Gray} 79.25 & \cellcolor{Gray} 71.17  \\ 
                                    & \cellcolor{Gray} MaMe($\tau=0.9$) & \cellcolor{Gray} 512 & \cellcolor{Gray} 58.10 & \cellcolor{Gray} 74.50  \\ \midrule
        \multirow{5}{*}{SigLIP2 (ViT-B/16)} 
                                    & Baseline                & 512        & 43.90        & 78.37       \\
                                    & ToMe(r=32)              & 512        & 50.89        & 76.46       \\ 
                                    & ToMe(r=64)              & 512        & 68.07        & 71.60       \\ 
                                    & \cellcolor{Gray} MaMe($\tau=0.9$) & \cellcolor{Gray} 512 & \cellcolor{Gray} 76.15  & \cellcolor{Gray} 75.09  \\
                                    & \cellcolor{Gray} MaMe($\tau=0.95$) & \cellcolor{Gray} 512 & \cellcolor{Gray} 56.15  & \cellcolor{Gray} 78.02  \\ 

        \bottomrule 
        \end{tabular}
    }

\end{table}

\paragraph{Text-Image to Text}
We investigated the impact of MaMe on the LLaVA-1.5-7B model\cite{liu2023llava} with the VLMEvalKit framework~\cite{duan2024vlmevalkit}, evaluating performance on COCO Caption\cite{lin2014coco} task. The token merging were applied to the visual encoder, aiming to \textbf{reduce the number of visual tokens fed into the large language model}, thereby enhancing computational efficiency. We compare the baseline LLaVA-1.5-7B against two distinct token merging strategies: ToMe, employing a fixed reduction number of $r=8$ per layer, and MaMe, which utilizes a similarity threshold of $\tau=0.8$.

\begin{table}[htbp!] 
\centering
\caption{COCO caption task performance of the LLaVA-1.5-7B model and its modified versions by ToMe and MaMe.}
\label{tab:coco_caption_results}
\renewcommand\arraystretch{1.3}
\resizebox{\columnwidth}{!}{ 
\begin{tabular}{lccccccc}
\toprule
\textbf{Method} & \textbf{Latency(s)} &\textbf{Bleu-1} & \textbf{Bleu-2} & \textbf{Bleu-3} & \textbf{Bleu-4} & \textbf{ROUGE\_L} & \textbf{CIDEr} \\
\midrule
LLaVA-1.5-7B & 3.12 & 20.72 & 13.28 & 8.08 & 4.93 & 20.94 & 0.71 \\
+ ToMe & 2.30 &20.15 & 12.90 & 7.90 & 4.89 & 21.67 & 1.60 \\
\rowcolor[gray]{0.9}
+ MaMe & 2.61 & 20.10 & 12.87 & 7.87 & 4.83 & 21.69 & \textbf{2.71} \\
\bottomrule
\end{tabular}
}

\end{table}

\begin{table}[!ht] 
\centering  
\small

\begin{tabular}{l p{0.8\columnwidth} }  
\toprule
 \multicolumn{2}{l}{\bf \small COCO Caption Task}  \\
\midrule
&  \includegraphics[width=0.3\columnwidth]{fig/2834.png} \\ 

\small \textbf{Question:} &  \small Please describe this image in general. Directly provide the description, do not include prefix like "This image depicts".  \\
 \small \textbf{LLaVA:} &  \small The image features a white plate with a variety of food items on it. There are two pieces of broccoli, one on the left side and the other on the right side of the plate. Additionally, there are two pieces of fish, one on the left side and the other on the right side of the plate. A fork is placed on the right side of the plate, ready to be used for eating. The plate is placed on a dining table, which is the main focus of the image. \\

 \small \textbf{+ToMe:} &  \small The image features a white plate with a variety of food items on it. There are two pieces of fish, one on the left side and the other on the right side of the plate. A generous portion of broccoli is spread across the plate, with some pieces located near the fish and others scattered around the plate. A fork and a knife are placed on the table, ready for use. The dining table occupies most of the image, with the plate and utensils placed on it. \\

 \small \textbf{+MaMe:} &  \small The image features a white plate with a variety of food items on it. There are two pieces of broccoli, one on the left side and the other on the right side of the plate. Two pieces of fish are also present on the plate, located \textcolor{blue}{towards the center}. 

 \small In addition to the food, there are utensils on the table. A fork is placed \textcolor{blue}{on the left} side of the plate, while a knife is located \textcolor{blue}{on the right} side. A \textcolor{green}{cup} can be seen \textcolor{blue}{on the left side of the table}, and a \textcolor{red}{bowl} is placed on the right side. The dining table occupies the majority of the image, with the plate and utensils placed \textcolor{blue}{on top of it}.
 \\
\bottomrule
\end{tabular}

\caption{MaMe correctly identified core components (plate, broccoli, fish, fork, knife) and their spatial arrangements, plus additional items (cup) that the other models miss. While it includes an erroneous "bowl," its output is structured with greater descriptive precision compared to the more generic and less detailed descriptions from LLaVA and ToMe. More results are in the supplementary material.} 
\label{tab:coco_caption} 
\end{table}

\paragraph{Results}  
Table~\ref{tab:coco_caption_results} shows MaMe achieved a CIDEr score of 2.71, a 3.8× improvement over the baseline and 69\% higher than ToMe (1.60)—while maintaining competitive Bleu and ROUGE\_L scores. Qualitative results in Table~\ref{tab:coco_caption} further confirm MaMe’s superiority, with more accurate object identification and spatial descriptions, aligning better with human consensus. These findings demonstrate that MaMe enhances both efficiency and caption quality.

\subsection{Video Classification}
\paragraph{Setup}  We apply token merging to VideoMAE\cite{tong2022videomae} models' vision encoder and compare MaMe with ToMe on Kinetics-400 validation set\cite{kay2017kinetics}.  For evaluation, 16 frames were sampled per video clip, each resized to a $224\times224$ resolution. We report Top-1 accuracy and inference throughput, measured in videos per second (videos/s) on a 3090 GPU with FP16 precision. 

\paragraph{Results}  Table \ref{tab:videomae} shows token merging accelerates video transformer inference. For VideoMAE-B, the baseline achieved 76.81\% accuracy with 13.24 videos/s throughput. MaMe($\tau=0.9$) maintained 76.03\% accuracy while reaching 13.33 videos/s, outperforming ToMe(r=128) at 14.06 videos/s but 3.47\% lower accuracy. MaMe($\tau=0.85$) achieved 13.81 videos/s comparable to ToMe(r=96). With VideoMAE-L, the baseline achieved 82.31\% accuracy at 6.25 videos/s. While ToMe(r=32) increased throughput to 6.97 videos/s, MaMe($\tau=0.8$) reached 9.28 videos/s with 81.47\% accuracy, showing a 49\% speed increase. This demonstrates MaMe's effectiveness for larger models in balancing throughput gains with performance. 

\begin{table}[ht!]
\small
\centering
\caption{Results of VideoMAE on action recognition benchmarks. Inference throughput is measured on a 3090 GPU with FP16 precision.}
\label{tab:videomae}
\renewcommand\arraystretch{1.5}
\resizebox{\columnwidth}{!}{
        \begin{tabular}{ccccc}
        \toprule
        \multirow{2}{*}{Model} & \multirow{2}{*}{Method} & Input  & Throughput & Top-1 Acc \\
                               &                         & (FxHW)   & (videos/s) & (\%) \\ \midrule
        \multicolumn{5}{c}{\textbf{Action Recognition on Kinetics-400}} \\ \midrule
        \multirow{5}{*}{\shortstack{VideoMAE-B}}
                               & Baseline                & 16x224  & 13.24 & 76.81 \\
                               & ToMe(r=96)              & 16x224  & 13.76 & 75.54 \\
                               & ToMe(r=128)             & 16x224  & 14.06 & 73.34 \\
                               & \cellcolor{Gray} MaMe($\tau=0.8$) & \cellcolor{Gray} 16x224 &  \cellcolor{Gray} 13.81 & \cellcolor{Gray} 74.23 \\ 
                               & \cellcolor{Gray} MaMe($\tau=0.9$) & \cellcolor{Gray} 16x224 &  \cellcolor{Gray} 13.33 & \cellcolor{Gray} 76.03 \\ \midrule
        \multirow{3}{*}{\shortstack{VideoMAE-L}}
                               & Baseline                & 16x224  & 6.25  & 82.31 \\
                               & ToMe(r=32)             & 16x224  & 6.97  & 82.05 \\
                               & \cellcolor{Gray} MaMe($\tau=0.8$) & \cellcolor{Gray} 16x224  & \cellcolor{Gray} 9.28 & \cellcolor{Gray} 81.47 \\ 

        \bottomrule
        \end{tabular}
        }

\end{table}

\begin{figure*}[t!]
\centering
\includegraphics[width=0.48\linewidth]{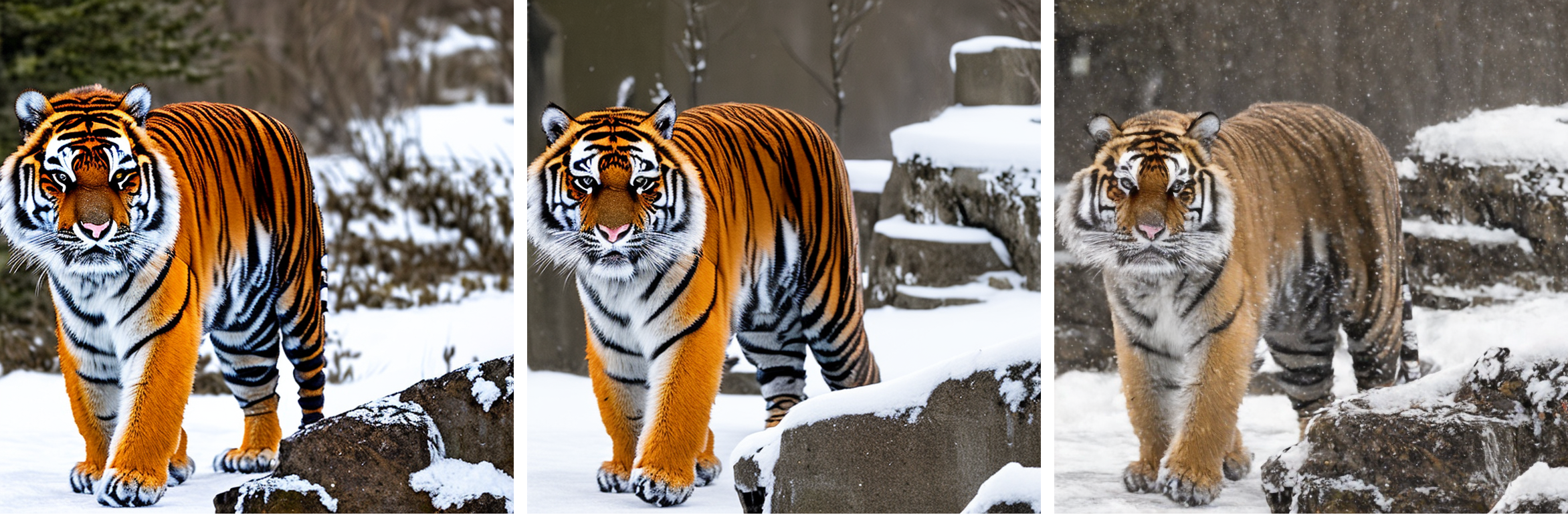}%
\hfill
\includegraphics[width=0.48\linewidth]{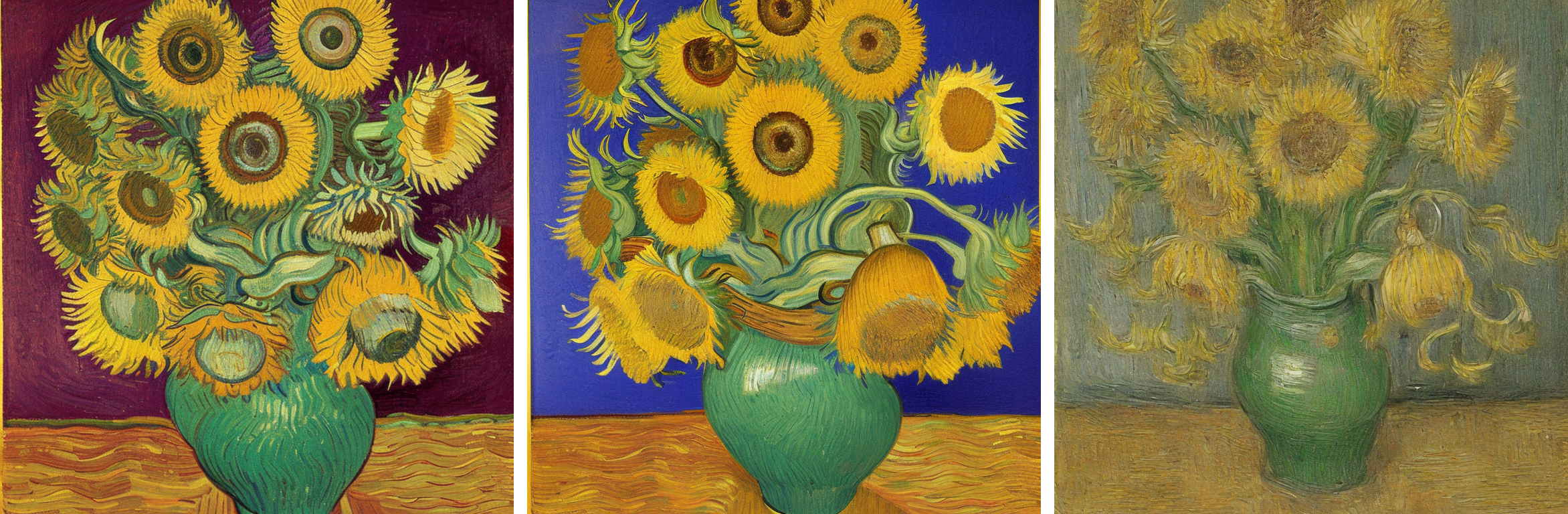}%
\\
\includegraphics[width=0.48\linewidth]{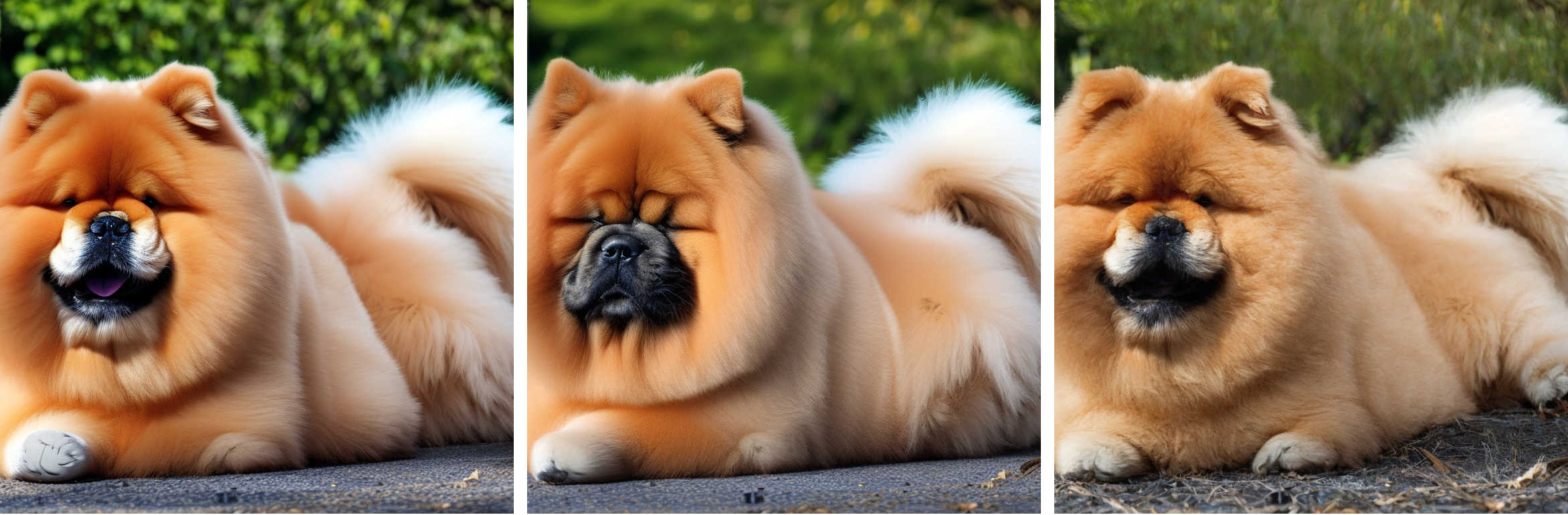}
\hfill
\includegraphics[width=0.48\linewidth]{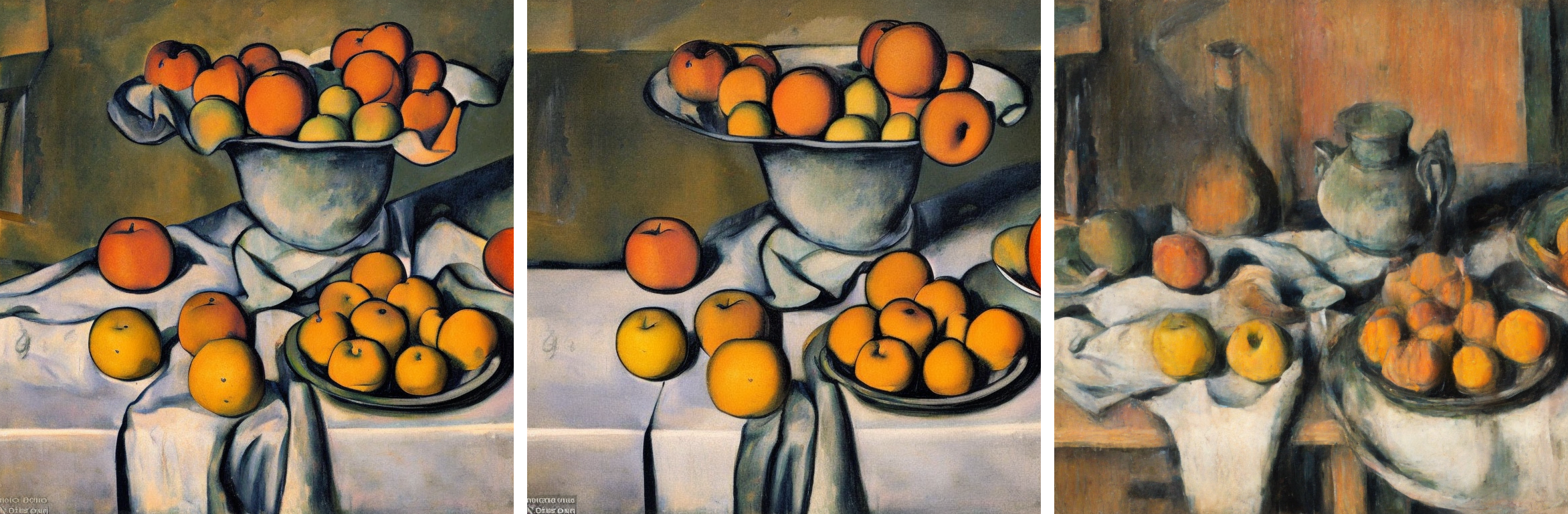}

\vspace{-5pt}
\caption{\textbf{Qualitative Results.} Images from left to right generated by Stable Diffusion v2.1, ToMeSD and MaMe+MaRe. MaMe+MaRe retains high-frequency details (left) and produce images with higher aesthetic quality (right). More results are in the supplementary material.}
\label{fig:hair-animals}
\end{figure*}

\begin{figure*}[t!]
    \centering
    \setlength{\tabcolsep}{0.1pt} 
    \begin{tabular}{ccccc}

        \includegraphics[width=0.198\linewidth]{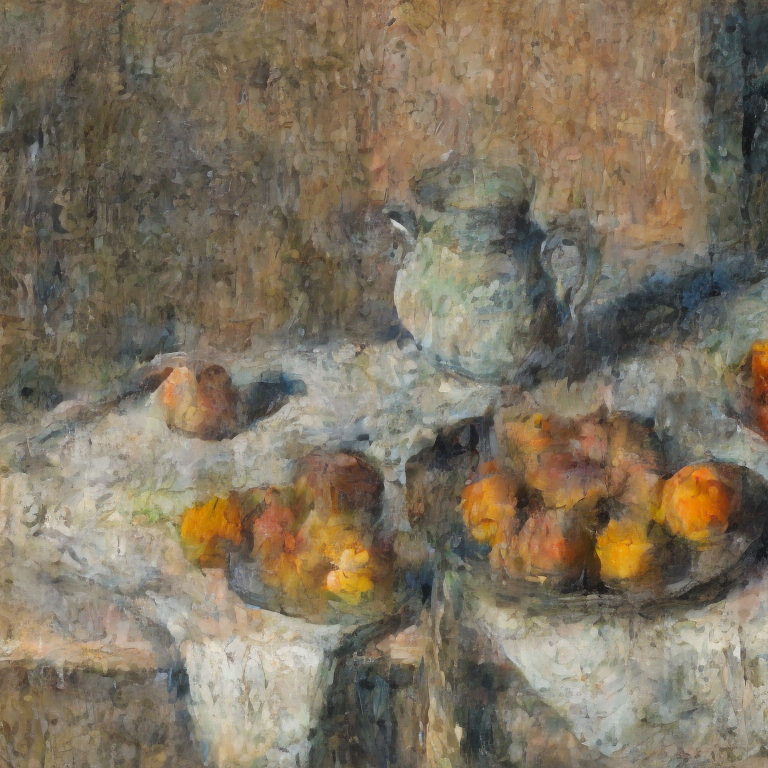} &
        \includegraphics[width=0.198\linewidth]{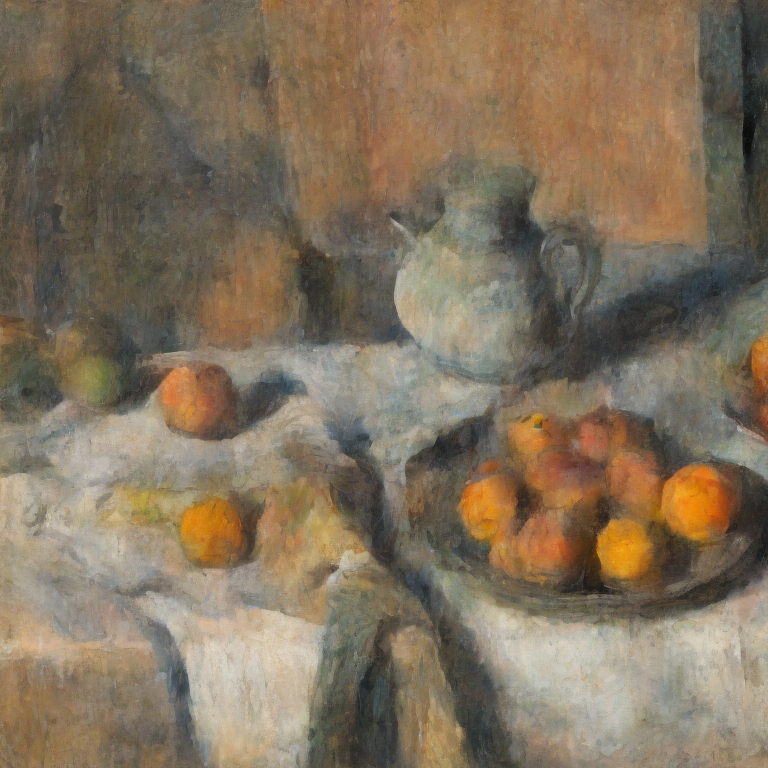} &
        \includegraphics[width=0.198\linewidth]{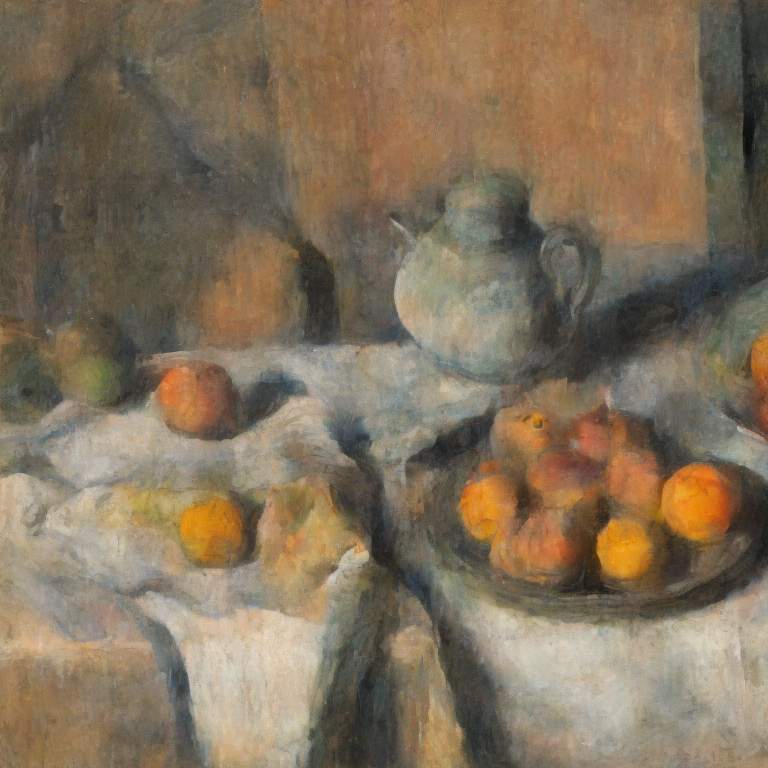} &
        \includegraphics[width=0.198\linewidth]{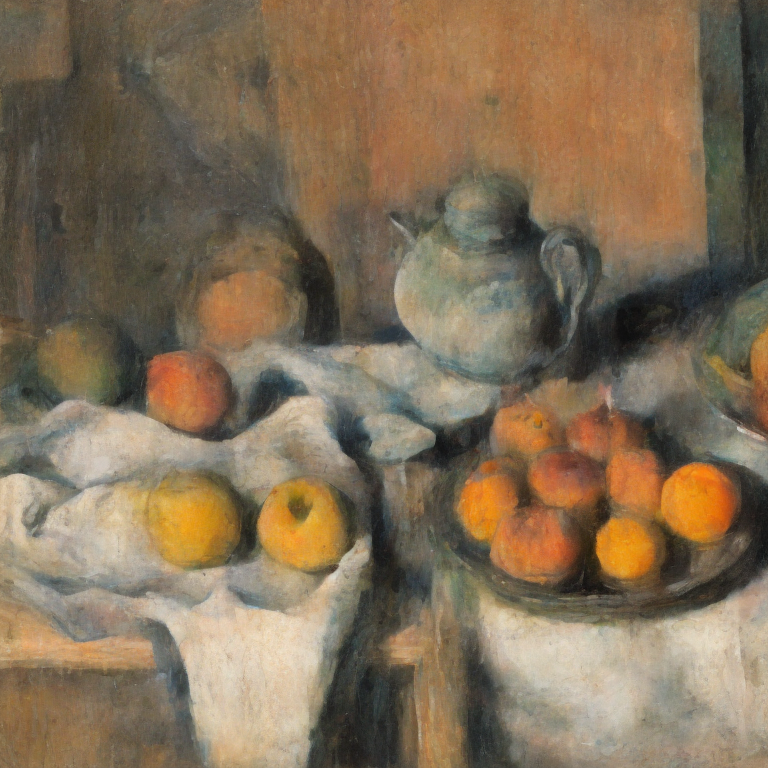} &
        \includegraphics[width=0.198\linewidth]{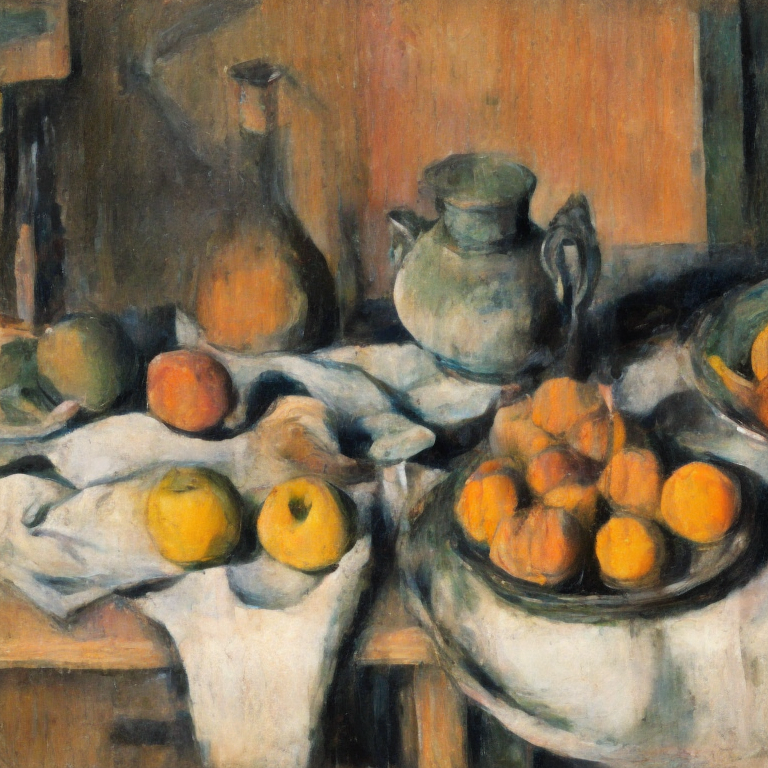} \\

    \end{tabular}
    \vspace{-5pt}
    \caption{\textbf{Qualitative Results.} The image generated by the model with MaMe. The clarity of the picture changed with the similarity threshold. A higher similarity threshold brings a more detailed and clearer painting. More results are in the supplementary material.}
    \label{fig:imgen_impression}
\end{figure*}

\subsection{Image Synthesis}
\paragraph{Setup}
We employed a baseline (unmodified Stable Diffusion v2.1\cite{Rombach_2022_CVPR} ) and modified models with two token merging methods: ToMeSD~\cite{bolya2023tome_sd} and proposed MaMe+MaRe, to generate 2,000 images at a resolution of $768 \times 768$ with the DPMSolver sampler, seed 42, a guidance scale of 7.5 and FP32 precision. The prompts were created from 2,000 randomly selected from the GEMRec dataset~\cite{GEMRec} and ImageNet–1K names of the classes. According to the formula~\ref{eq:gen}, token merging and restore were applied before and after the attention calculation. For ToMeSD, we examined two configurations: removing 25\% and 50\% of the tokens. For MaMe+MaRe, we tested two similarity thresholds: $\tau=0.7$ and $\tau=0.8$. To evaluate speed, we averaged the time taken to generate images across all 2,000 samples on a 3090 GPU.

\paragraph{Results}
The performance of baseline Stable Diffusion v2.1 and its variants, modified by ToMeSD and MaMe, is summarized in Table \ref{tab:text_to_image}. The evaluation focuses on two aspects: inference speed, measured by latency (seconds per image generation), and image quality, assessed through a suite of metrics including FID (Fréchet Inception Distance)\cite{heusel2017gans}, IS (Inception Score)\cite{salimans2016improved}, LPIPS (Learned Perceptual Image Patch Similarity)\cite{zhang2018perceptual}, PSNR (Peak Signal-to-Noise Ratio)\cite{gonzalez2008digital}, SSIM (Structural Similarity Index Measure)\cite{wang2004image}, and CLIP Score\cite{radford2021learningtransferablevisualmodels}.

\begin{table}[ht]
\centering
\footnotesize
\caption{Text-to-image generation performance comparison results with SD2.1 (baseline) in generating 768×768 images, based on 50 sampling steps.} 
\label{tab:text_to_image} 
\renewcommand\arraystretch{1.2}
\tabcolsep=0.02cm 
\resizebox{\columnwidth}{!}{
\begin{tabular}{@{}lcccccccc@{}} 
\toprule
\multirow{2}{*}{} & \multicolumn{1}{c}{Speed} & \multicolumn{6}{c}{Image Quality} \\ 
\cmidrule(l){2-2} \cmidrule(l){3-8} 
                        & Latency (s) $\downarrow$ & FID $\downarrow$ & IS $\uparrow$ & LPIPS $\downarrow$ & PSNR $\uparrow$ & SSIM $\uparrow$ & CLIP $\uparrow$ \\ 
\midrule
baseline                & 17.26                        & 1.93             & 22.75         & 0.68              & 8.80            & 0.17            & 23.41          \\
ToMeSD(25\% ratio)              & 14.81                        & 2.04             & 23.47         & 0.64              & 9.07            & 0.19            & 23.48          \\
ToMeSD(50\% ratio)              & 12.67                        & 3.68             & 22.21         & 0.65              & 9.13            & 0.21            & 23.51          \\ 
\rowcolor[gray]{0.9}
MaMe+MaRe($\tau=0.7$)        & 11.87                        & 2.54             & 20.84         & 0.63              & 9.77            & 0.20            & 24.15          \\
\rowcolor[gray]{0.9}
MaMe+MaRe($\tau=0.8$)        & 12.45                        & 2.28             & 23.31         & 0.63              & 9.79            & 0.21            & 24.18          \\
\bottomrule
\end{tabular}
}

\end{table}

\begin{figure*}[t!] 
    \centering
    
    \subfloat[\normalfont \footnotesize \textbf{Feature Choice.} \label{tab:ablation_feature_choice} ]{
        \begin{minipage}{0.23\linewidth}
            \centering
            \tablestyle{4pt}{1.25}
            \begin{tabular}{x{26}x{22}x{22}} feature & acc & im/s \\ \hline
                \default{\texttt{x}} & \default{{\bf 83.35}} & \default{73.06} \\
                \texttt{k} & 71.63 & 72.45 \\
                \texttt{k-mean} & 69.01 & 75.06 \\
                & &\\
            \end{tabular}
        \end{minipage}
    }\hfill 
    \subfloat[\normalfont \footnotesize \textbf{Similarity Func.} \label{tab:ablation_distance_function} ]{
        \begin{minipage}{0.23\linewidth}
            \centering
            \tablestyle{4pt}{1.25}
            \begin{tabular}{y{32}x{22}x{22}} function & acc & im/s \\ \hline
                eucl & 81.58 & 73.60 \\
                \default{cosine} & \default{ 83.35} & \default{{73.06}} \\
                dot & 80.43 & 81.14 \\
                softmax & 61.90 & 80.33\\
            \end{tabular}
        \end{minipage}
    }\hfill 
    \subfloat[ \normalfont \footnotesize \textbf{Partition Style.} \label{tab:ablation_partition} ]{
        \begin{minipage}{0.23\linewidth}
            \centering
            \tablestyle{4pt}{1.25}
            \begin{tabular}{y{38}x{22}x{22}} order & acc & im/s \\ \hline
                sequential & 84.14 & {71.81} \\
                \default{alternating} & \default{83.35} & \default{73.06}\\
                random & 83.24 & 73.74 \\
                & & \\
            \end{tabular}
        \end{minipage}
    }\hfill 
    \subfloat[ \normalfont \footnotesize \textbf{Refine Weight.} \label{tab:ablation_adaptive} ]{
        \begin{minipage}{0.23\linewidth}
            \centering
            \tablestyle{4pt}{1.25} 
            \begin{tabular}{y{18}x{18}x{20}x{20}}
               src & refine & acc & im/s \\ 
               \shline
               mae & & 77.51& 85.59\\
               mae & \checkmark & \default{80.02} & \default{78.96}\\
               \shline
               aug & & 72.47 & 78.59 \\ 
               aug & \checkmark & \default{83.35} & \default{73.06} \\
            \end{tabular}
        \end{minipage}
    }

    \vspace{-5pt}
    \caption{Ablation experiments using AugReg ViT-B/16. Our default settings are marked in Gray. We report Top-1 accuracy (acc) and model inference throughput (im/s) with FP32 precision and batchsize 1 on a 3090 GPU. The visualization results of different methods are shown in Figure~\ref{fig:vis_ablation}.}
    \label{tab:ablation_results}

    \vspace{1em} 

    \includegraphics[width=\textwidth]{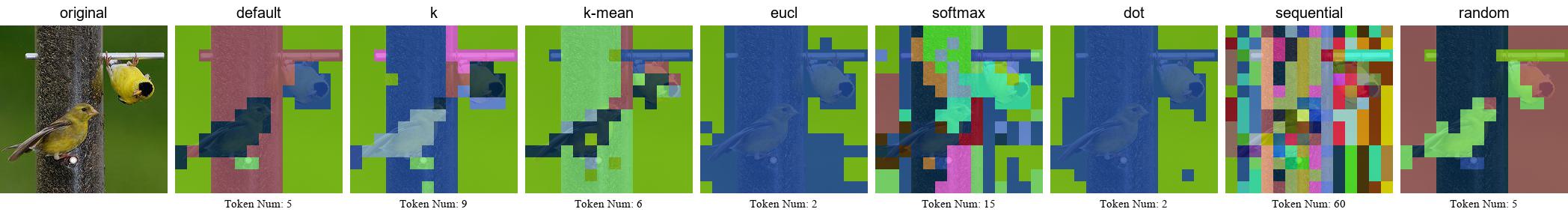}\\
    \vspace{2pt} 
     \includegraphics[width=\textwidth]{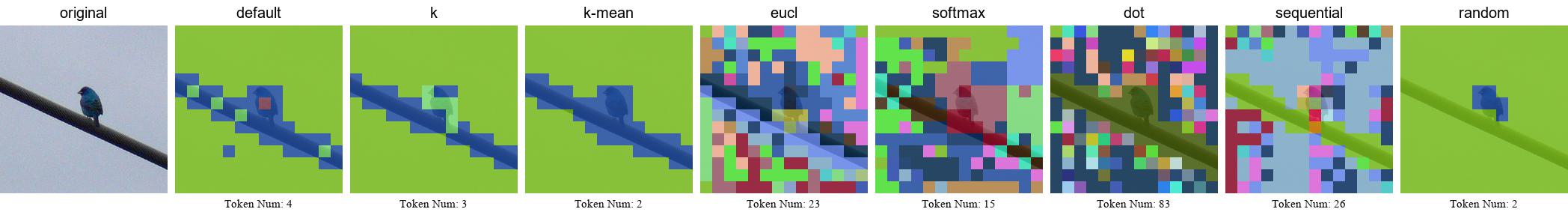}
    
    \vspace{-5pt}
    \caption{The visualization of the final token count in the 8th block of the AugReg ViT-B/16. Each color square represents a distinct type of token.}
    \label{fig:vis_ablation}
    
\end{figure*}

Table~\ref{tab:text_to_image} shows MaMe+MaRe achieves highest acceleration, reducing latency from 17.26s to 11.87s at $\tau=0.7$ (31.2\% speedup) and 12.45s at $\tau=0.8$ (27.9\% speedup), surpassing ToMeSD's 14.81s (25\% ratio) and 12.67s (50\% ratio). MaMe+MaRe delivers better image fidelity with lowest LPIPS scores (0.63) and highest PSNR values (9.77--9.79), indicating superior quality than baseline and ToMeSD variants. Semantic alignment, measured by CLIP Score, further distinguishes MaMe+MaRe. At $\tau=0.8$, MaMe+MaRe reaches 24.18, exceeding the baseline (23.41) by 3.3\% and ToMeSD(50\%) (23.51) by 2.8\%. This suggests MaMe+MaRe pipeline preserves text-image semantic consistency better. While ToMeSD(50\%) achieves comparable speed (12.67s), it has higher FID degradation (3.68 vs. 1.93 baseline), whereas MaMe+MaRe($\tau=0.8$) maintains a better FID of 2.28. And we discovered that image clarity can be controlled by adjusting the similarity threshold as shown in Figure~\ref{fig:imgen_impression}, which demonstrates pipeline's unique tunability. These results show MaMe+MaRe surpasses the typical efficiency-quality trade-off, delivering better latency and quality for text-to-image generation.

\paragraph{Why does MaMe+MaRe enhance image quality?}\\
We think MaMe plays the role of a \textbf{high-pass filter} by adaptively compressing redundant, low-frequency information (e.g., smooth image regions), while preserving unique, high-frequency tokens (e.g., edges and textures) unchanged through into attention layers. And then the reduction of token count happens to mitigate the "attention dilution" effect, also known as “rank-collapse” or “token uniformity,”\cite{dong2021attention} where a large softmax denominator in MSA disperses finite attention capacity, suppressing significant tokens. Unique tokens, previously diluted, now compete within a smaller candidate set, thereby receiving higher attention scores, which means \textbf{amplifying high-frequency signals.} The preserved and enhanced details improve visual responses, also explaining the CIDEr score improvement on the COCO caption task and indicating MaMe's potential for enhancing the Vision Language Model (VLM). Finally, MaRe accurately restores the processed information back to the original spatial structure by maintaining the original token relationships. So the final result has benefited from a more focused attention computation while fully retaining its original spatial resolution and high-frequency detailed content.

\subsection{Ablation Study}

\subsubsection{Algorithmic Design Choices}

Our ablation studies extend to the core algorithmic components of MaMe, with results summarized in Table~\ref{tab:ablation_results}. Four key design dimensions are investigated:

\textbf{Feature Choice:} The raw token matrix \texttt{x} achieves optimal accuracy (83.35\%) with competitive throughput. Features \texttt{k} and \texttt{k-mean} show lower accuracy (71.63\% and 69.01\%), confirming \texttt{x} preserves the most discriminative information for merging decisions.

\textbf{Similarity Function:} Cosine similarity achieves the best balance (83.35\% accuracy, 73.06 im/s throughput). Dot product improves throughput (81.14 im/s) but sacrifices accuracy (80.43\%), while Euclidean and softmax-based methods underperform in either metric.

\textbf{Partition Style:} Sequential ordering maximizes accuracy (84.14\%) but reduces throughput (71.81 im/s). Alternating order offers the best trade-off (83.35\% accuracy, 73.06 im/s), outperforming random ordering, which slightly boosts throughput at the cost of accuracy.

\textbf{Adaptive Weight Refining:} AugReg models require pruning to achieve best 83.35\% accuracy. MAE models show moderate gains but remain less dependent on pruning.

\begin{figure*}[ht!]
\centering
\subfloat[ \normalfont \footnotesize Accuracy vs. Sim. Threshold with Alternating partition]{\includegraphics[width=3.2in]{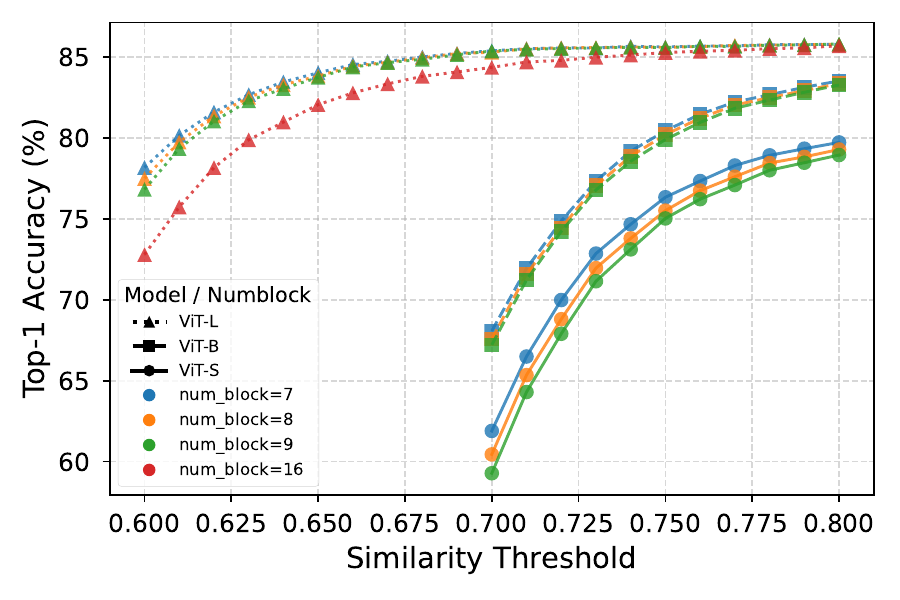}%
\label{fig_first_case}}
\hfil
\subfloat[ \normalfont \footnotesize Throughput vs. Sim. Threshold with Alternating partition]{\includegraphics[width=3.2in]{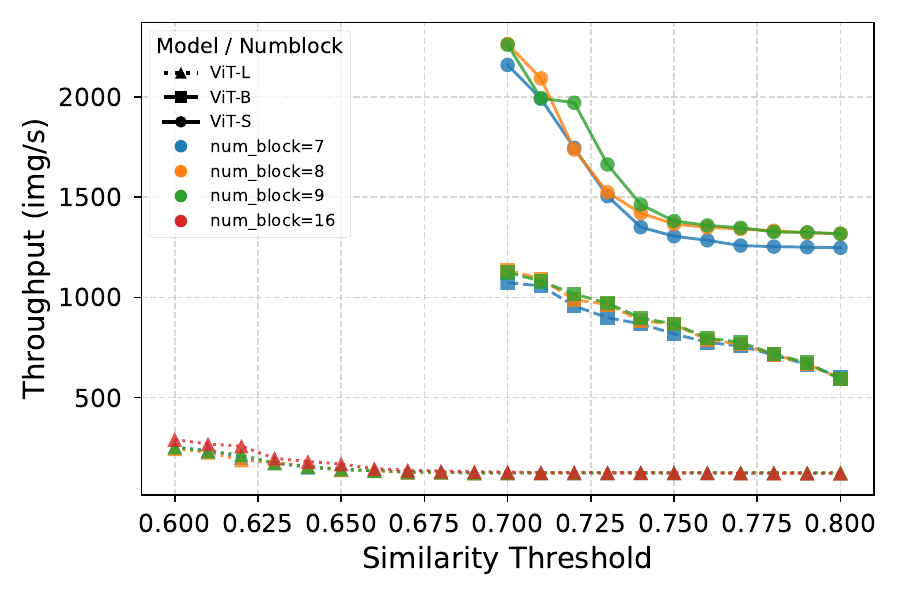}
\label{fig_second_case}}
\hfil
\subfloat[ \normalfont \footnotesize Accuracy vs. Sim. Threshold with Random partition]{\includegraphics[width=3.2in]{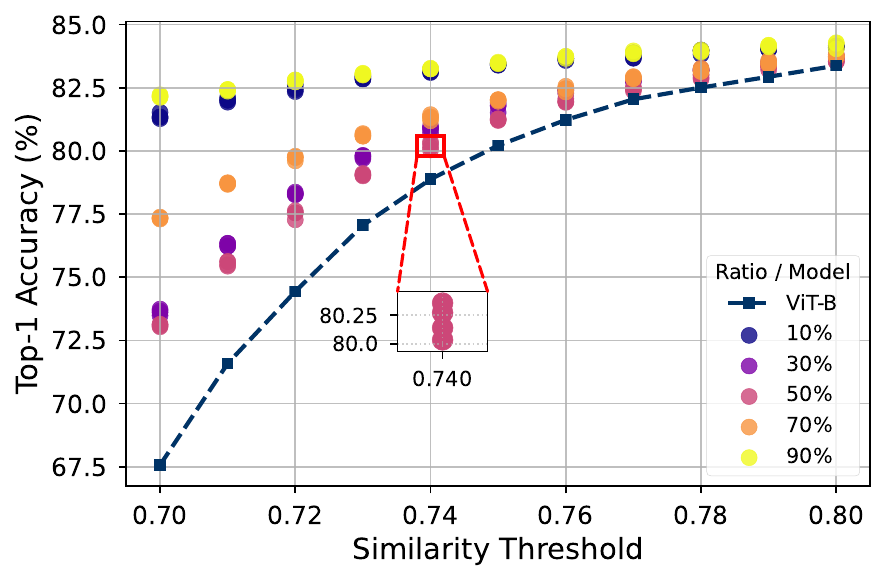}%
\label{fig_third_case}}
\hfil
\subfloat[ \normalfont \footnotesize Throughput vs. Sim. Threshold with Random partition]{\includegraphics[width=3.2in]{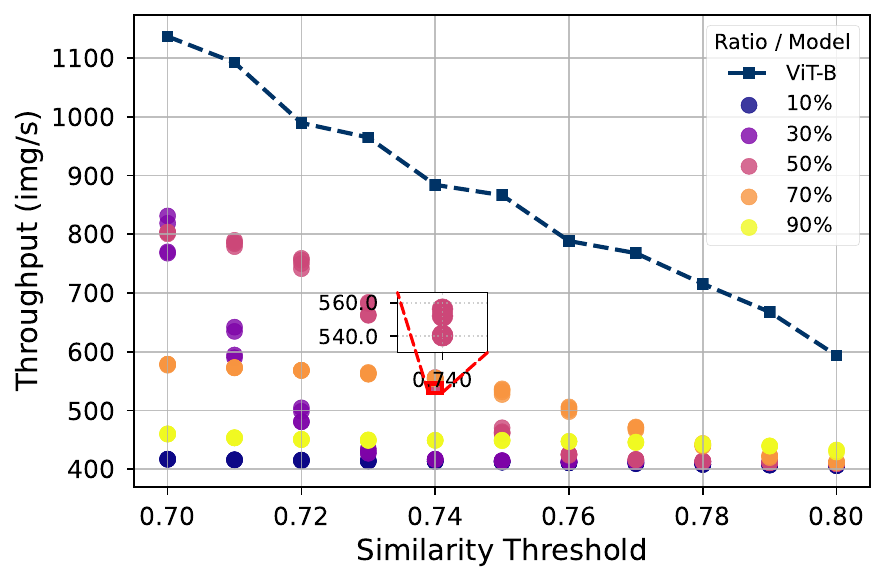}
\label{fig_fourth_case}}

\caption{The accuracy and throughput change with the similarity threshold under both alternating and random partitions. $num\_block$ denotes the number of blocks used for merging. There is a non-linear, saturating relationship between the similarity threshold and accuracy across ViT architectures. Five different random seeds are employed to conduct five experiments under different ratios of tokens as source tokens as shown in the box in the Figures (c) and (d), illustrating that the default, determined alternating partition curves represent a Pareto frontier.}
\label{fig:ablations}

\end{figure*}
\subsubsection{Where and What}

To investigate where MaMe should be applied within models and what similarity threshold yields optimal performance, we examine the joint impact of similarity threshold ($\tau$) and the number of blocks applying token merging ($num\_block$) on accuracy and throughput on ViT models as shown in Figure~\ref{fig:ablations}.

\textbf{Accuracy}
The relationship between similarity threshold and accuracy shows a non-linear pattern across ViT architectures, influenced by MaMe applied block depth. As the threshold increases from 0.6 to 0.8, accuracy improves rapidly before plateauing, indicating diminishing returns from stricter token retention and suggesting that exceeding a critical threshold sufficiently distinguishes features.

\textbf{Throughput}
Throughput monotonically decreases with similarity threshold, dropping sharply at low thresholds due to rising computational costs. Threshold sensitivity inversely correlates with model size: ViT-S experiences the steepest decline, followed by ViT-B and ViT-L, indicating larger models better mitigate merging overhead.

\textbf{Model-Scale Sensitivity}
Model sensitivity to token merging varies by size: ViT-L maintains $>$85\% accuracy across thresholds (0.6–0.8) with throughput gains, ViT-B shows moderate sensitivity, and ViT-S is most vulnerable (accuracy drops from ~79\% to ~60\% with aggressive merging). Larger models exhibit greater representational redundancy, allowing coarser merging with minimal performance loss.

\textbf{Random Partition}
The comparison between deterministic default configurations (dashed lines) and stochastic trials (scatter points) indicates that the default settings define a Pareto frontier: stochastic partitions yield higher accuracy (points above the dashed line) but lower throughput (points below the dashed line). This trade-off suggests that while stochasticity enhances accuracy, it undermines computational efficiency. The deterministic, alternating partition thus serves as a robust baseline, balancing performance and efficiency.

\section{Discussion}

\subsection{Pros and Cons}
One of the primary advantages of MaMe is its non-intrusive nature with respect to standard attention mechanisms. Unlike ToMe, which requires attention computation modifications, MaMe changed nothing and works seamlessly with optimized implementations like Flash Attention\cite{dao2023flashattention2}. Another advantage is MaMe's matrix-only operations are GPU-friendly, avoiding sorting or scattered writes that are GPU-inefficient. MaMe's drawback is that the optimal similarity threshold ($\tau$) for maximal effect require manual determination. Future work will develop automated methods for determining it.

\subsection{Future Work}
The observed improvements of MaMe on \textit{COCO Caption} and image generation quality suggest that MaMe effectively preserves semantic-rich features critical for generative tasks. This implies MaMe's potential in two areas: (1) \textit{Video understanding and generation}, where MaMe could optimize long-range spatiotemporal modeling by compressing redundant frame-level tokens; and (2) \textit{Vision-Language-Action (VLA) models}, where efficient token reduction may enhance real-time decision-making in embodied AI systems. 

Another frontier for MaMe lies in its application to LLMs. Token merging approaches like ToMe, which rely on sorting tokens by similarity, can break the causality constraint in autoregressive LLMs by allowing tokens to merge with future tokens. MaMe enables \textbf{causal token merging} by partitioning tokens into two mutually exclusive sets (destination tokens set for odd-indexed tokens like 1, 3, 5... and source tokens set for even-indexed tokens like 0, 2, 4, 6...) and applying a causality mask $M$ to set the upper triangular part of the final fusion weights $W_{ij}^{\text{F}}$ to zero by $W_{ij}^{\text{F}} \odot M_{ij}$, where $M_{i,j} = 1$ if $j\leq i$ and $0$ otherwise. This restricts each destination token (e.g., token 5) to merge only with preceding source tokens (e.g., tokens 0, 2, 4), preserving causality and allowing reduce KV cache size in LLMs.

\section{Conclusion}
\label{sec:con}
In this paper, we present MaMe and MaRe, forming a GPU-friendly plug-and-play token merging and restoration framework with pure matrix computations, eliminating the overhead of sorting or clustering. While MaMe, like ToMe, typically balances speed with performance trade-offs in most tasks, it achieves simultaneous improvements in both captioning and image synthesis, suggesting that MaMe reduces a multitude of mediocre tokens, allowing the importance of critical tokens to emerge during compression. We infer that MaMe can make performance better in more image and video understanding and generation tasks through direct application or fine-tuning.

\bibliographystyle{elsarticle-num}
\bibliography{ref}

\clearpage
\newpage
\appendix
  
\section{Appendix}
\begin{figure*}[!ht]
    \centering
    
    \includegraphics[width=\textwidth]{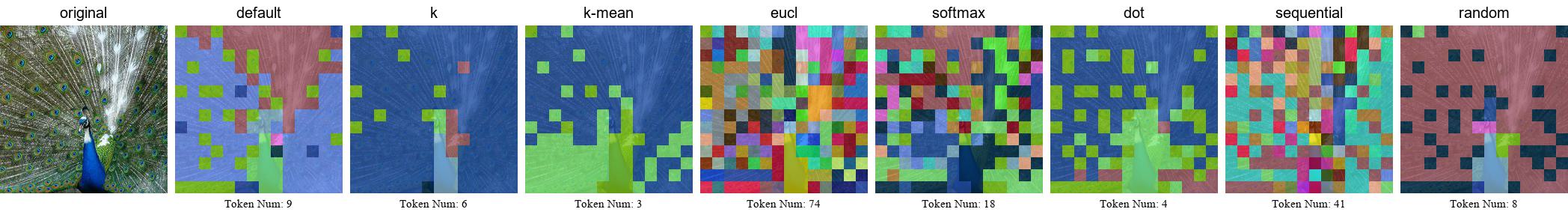}

    \includegraphics[width=\textwidth]{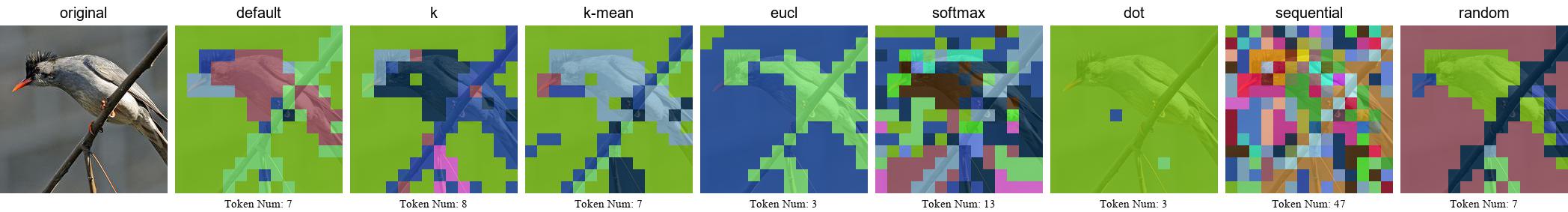}
    
    \includegraphics[width=\textwidth]{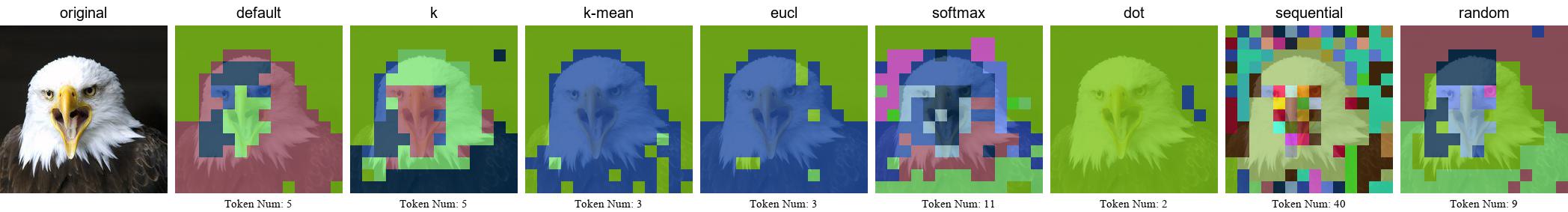}
    
    \includegraphics[width=\textwidth]{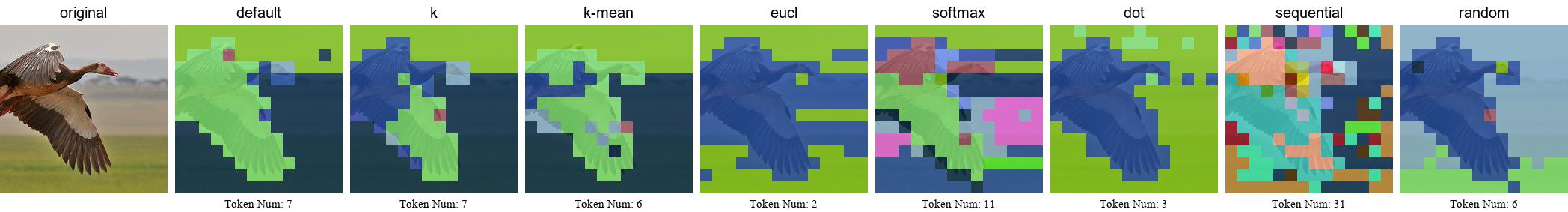}    

    \includegraphics[width=\textwidth]{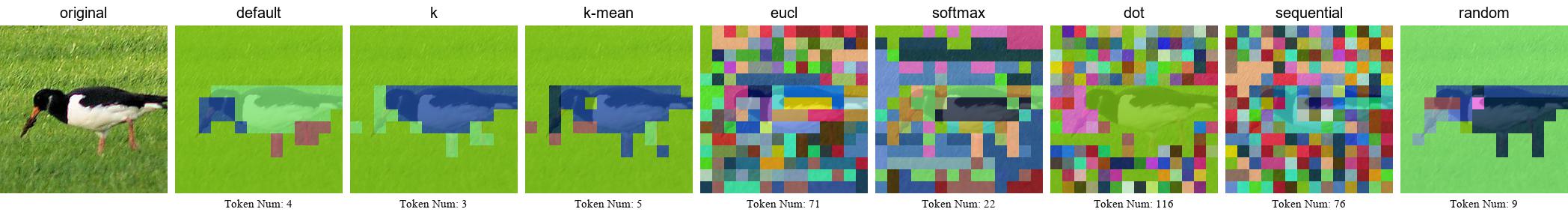}
    
    \includegraphics[width=\textwidth]{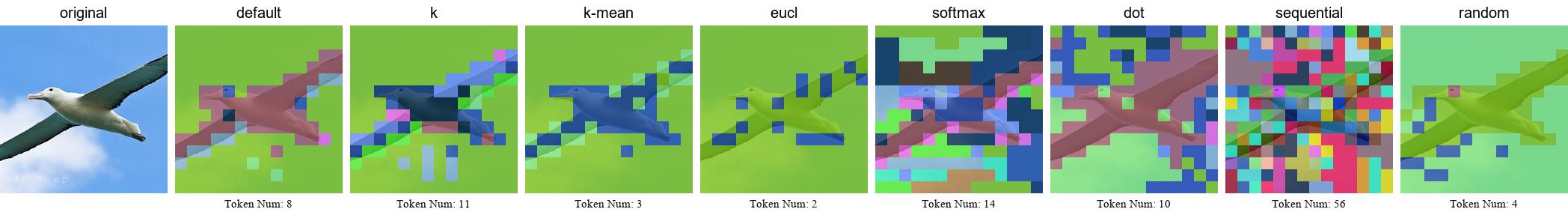}
    
    \includegraphics[width=\textwidth]{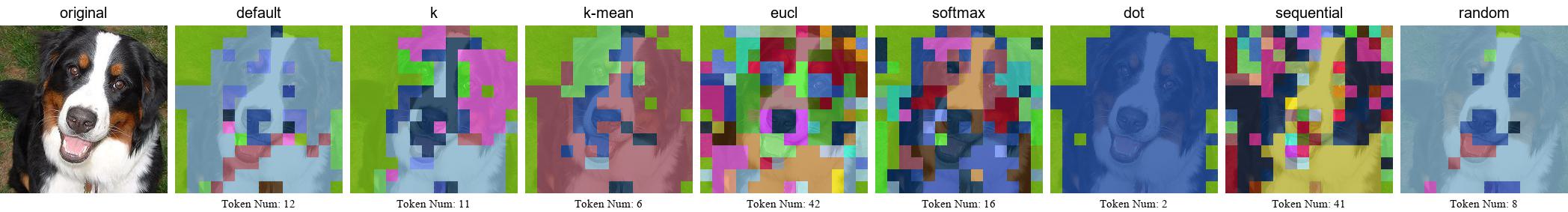}

    \includegraphics[width=\textwidth]{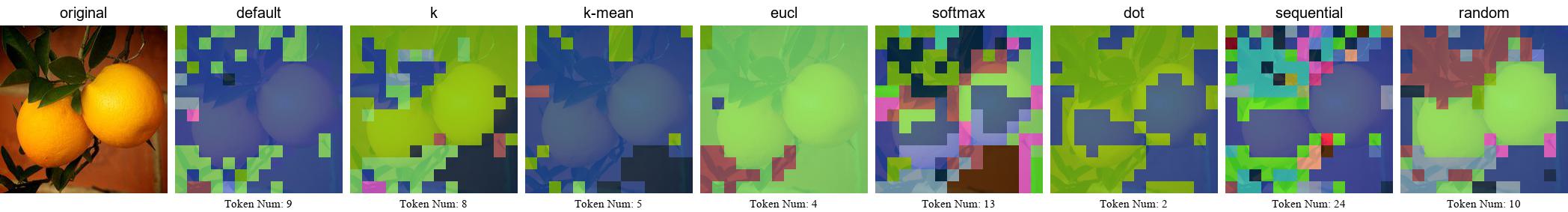}

    \includegraphics[width=\textwidth]{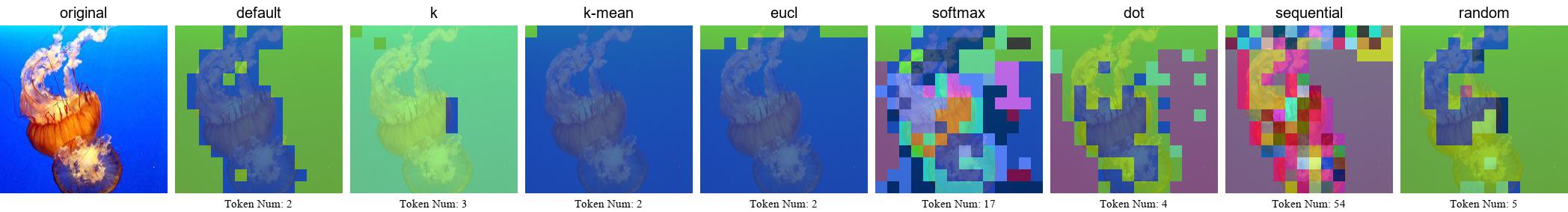}
    
    \vspace{-5pt}
    \caption{The visualization in the 8th block of the AugReg ViT-B/16 using MaMe with different settings. Each color square represents a distinct type of token. Default is our default method.}
\end{figure*}


\begin{figure*}[!ht]
\centering
\includegraphics[width=\textwidth]{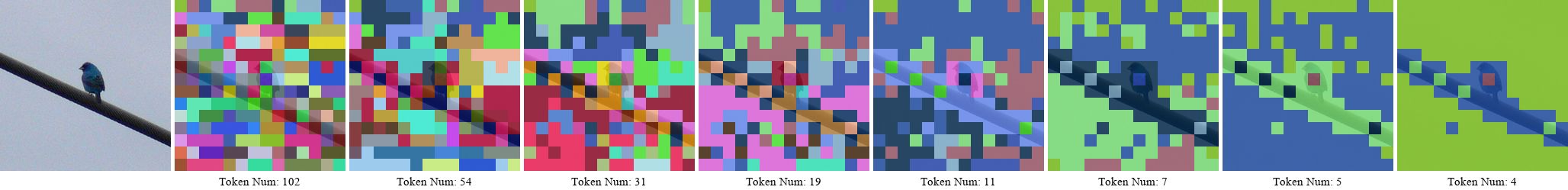}%
\\
\includegraphics[width=\textwidth]{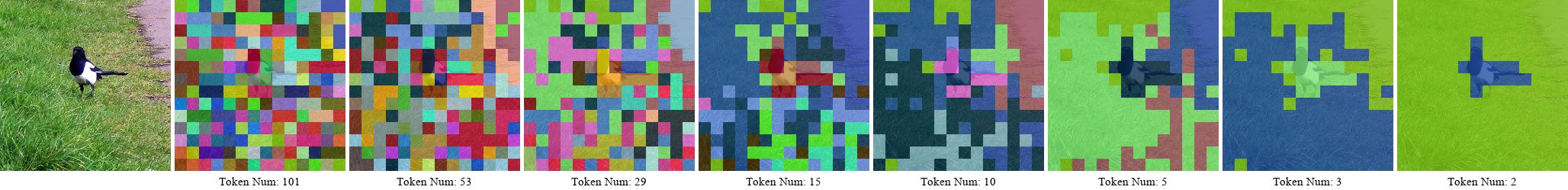}%
\\
\includegraphics[width=\textwidth]{fig/progress/n01818515_1466_x_8.jpg}%
\\
\includegraphics[width=\textwidth]{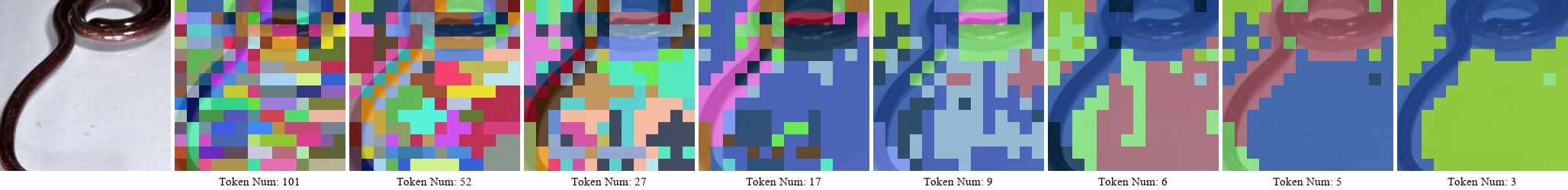}%
\\
\includegraphics[width=\textwidth]{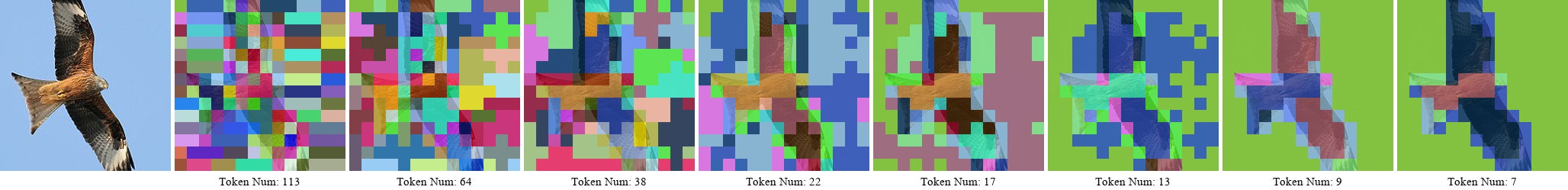}%
\\
\includegraphics[width=\textwidth]{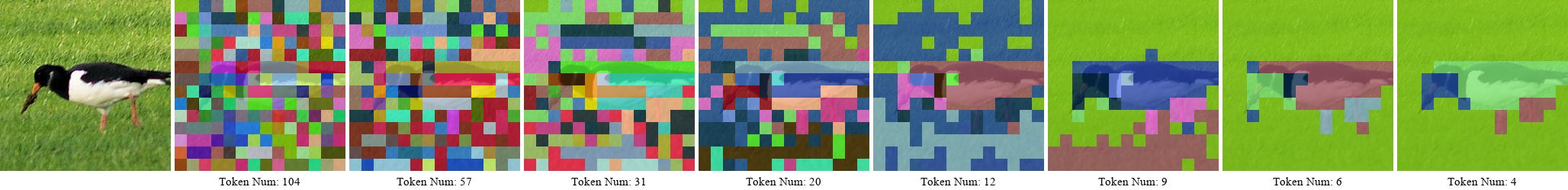}%
\\
\includegraphics[width=\textwidth]{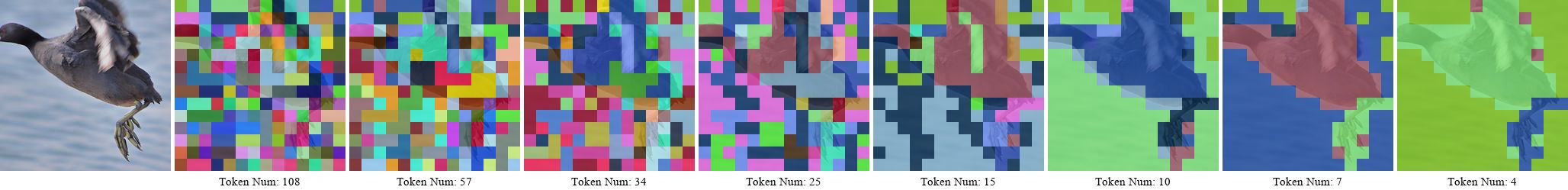}%
\\
\includegraphics[width=\textwidth]{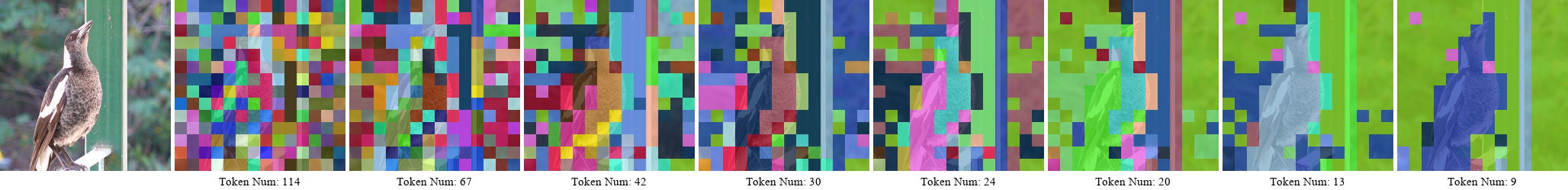}%
\\
\includegraphics[width=\textwidth]{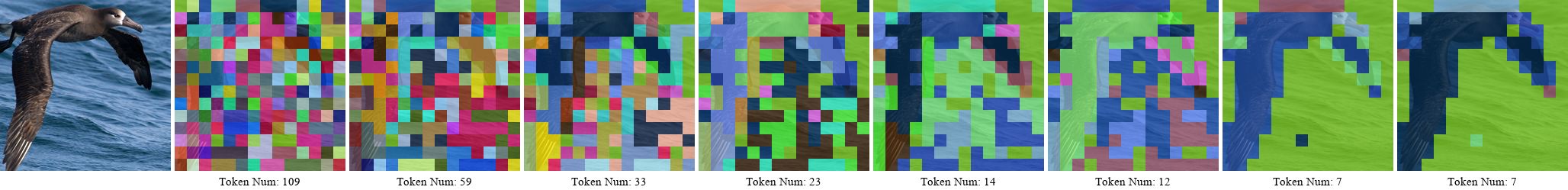} 
\\
\includegraphics[width=\textwidth]{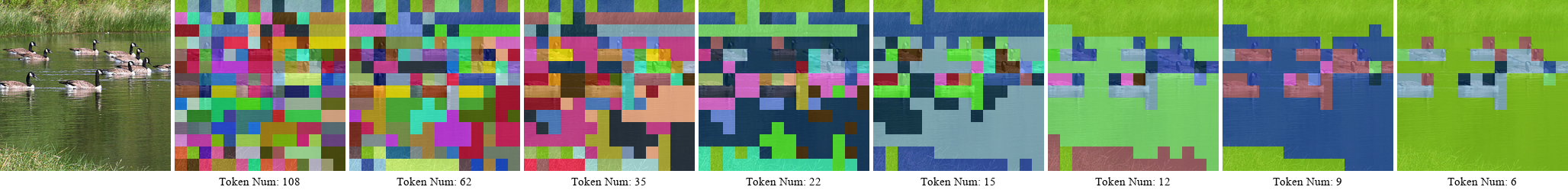}%
\\
\includegraphics[width=\textwidth]{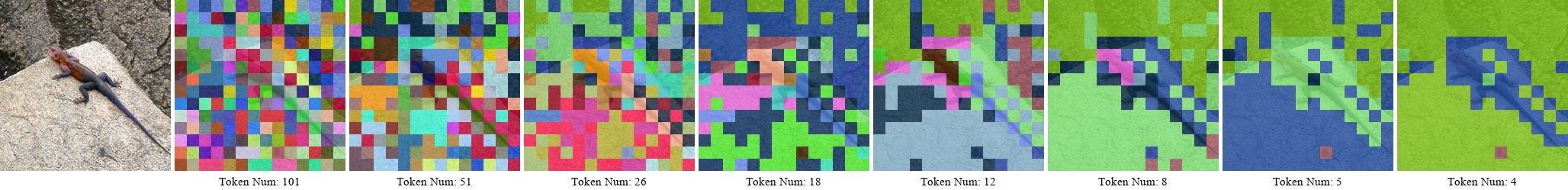}%

\vspace{-5pt}
\caption{The visualization illustrates the progression of token count reduction in the first 8 blocks of the AugReg ViT-B/16 with MaMe. Each color square represents a distinct type of token.}
\end{figure*}

\begin{table*}[!ht] 
\centering  
\small
\begin{tabular}{l p{0.8\textwidth} } 
\toprule
 \multicolumn{2}{l}{\bf \normalsize COCO Caption Task}  \\
\midrule
&  \includegraphics[width=0.3\columnwidth]{fig/2892.jpg} \\ 

\normalsize \textbf{Question:} &  \normalsize Please describe this image in general. Directly provide the description, do not include prefix like "This image depicts".  \\
 \normalsize \textbf{LLaVA:} &  \normalsize 2 people are standing in front of a table with a variety of wine bottles and a plate of food. The man is holding a wine bottle and the woman is holding a wine glass. They are both smiling and enjoying their time together. The table is filled with wine bottles and a few wine glasses, and there is a cake on the table as well. \\

 \normalsize \textbf{+ToMe:} &  \normalsize 2 people are standing at a table with a variety of wine bottles and glasses. The man is wearing a yellow shirt and the woman is wearing a black shirt. They are both holding wine glasses and appear to be tasting different wines. There are several bottles of wine on the table, some of which are placed closer to the man and others near the woman. A plate of food is also present on the table, adding to the ambiance of the wine tasting event.\\

 \normalsize \textbf{+MaMe:} &  \normalsize 2 people are standing at a table with a variety of food and drinks. The man is wearing a \textcolor{blue}{yellow shirt} and the woman is wearing a \textcolor{blue}{black shirt}. They are both holding wine glasses and appear to be enjoying a meal together. The table is filled with several \textcolor{blue}{bottles of wine, a plate of food, and a fork}. The scene is set in a restaurant with a \textcolor{blue}{window view}.

 \\
\bottomrule
\end{tabular}
\caption{MaMe provides the most accurate and comprehensive description, correctly identifying clothing details (yellow shirt, black shirt) and contextual elements (restaurant setting, window view). ToMe captures basic elements accurately but lacks environmental context. LLaVA omitting important visual details like clothing colors. MaMe demonstrates superior visual grounding and contextual understanding.} 
\label{tab:coco_caption} 
\end{table*}

\begin{table*}[!ht] 
\centering  
\small
\begin{tabular}{l p{0.8\textwidth} } 
\toprule
 \multicolumn{2}{l}{\bf \normalsize COCO Caption Task}  \\
\midrule
&  \includegraphics[width=0.3\columnwidth]{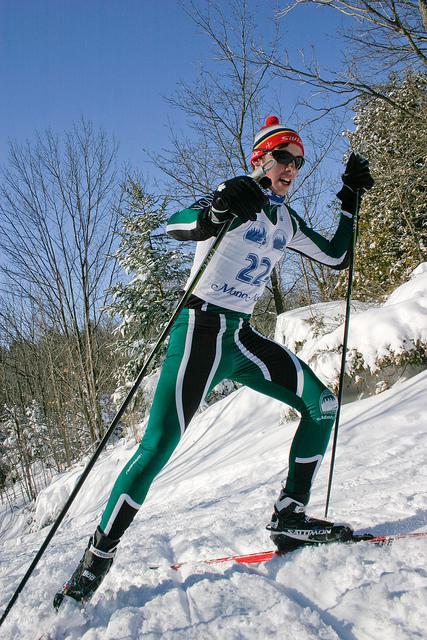} \\ 

\normalsize \textbf{Question:} &  \normalsize Please describe this image in general. Directly provide the description, do not include prefix like "This image depicts".  \\

 \normalsize \textbf{LLaVA:} &  \normalsize A man wearing a green and white ski suit is skiing down a snowy hill. He is holding ski poles in his hands and appears to be enjoying the activity. The man is the main focus of the image, and the snowy hill and trees in the background provide a picturesque winter scene. \\

 \normalsize \textbf{+ToMe:} &  \normalsize 24 is the number on the man's outfit.\\

 \normalsize \textbf{+MaMe:} &  \normalsize A man wearing a green and white ski suit is skiing down a snowy hill. He is holding ski poles in his hands and appears to be enjoying the activity. The man is wearing a \textcolor{blue}{hat and goggles, which are essential for protecting his eyes from the cold and snow}. The scene captures the excitement and thrill of skiing in a beautiful winter landscape.

 \\
\bottomrule
\end{tabular}
\caption{MaMe delivers the most comprehensive and context-aware description, accurately identifying all visual elements (ski suit, poles, hat, goggles) and providing relevant contextual reasoning about protective gear. LLaVA covers basics but omits key details like headgear, while ToMe fails entirely, producing only an unrelated fragment ("number 24").} 
\label{tab:coco_caption} 
\end{table*}

\begin{figure*}[!htbp]
\centering
\includegraphics[width=0.48\linewidth]{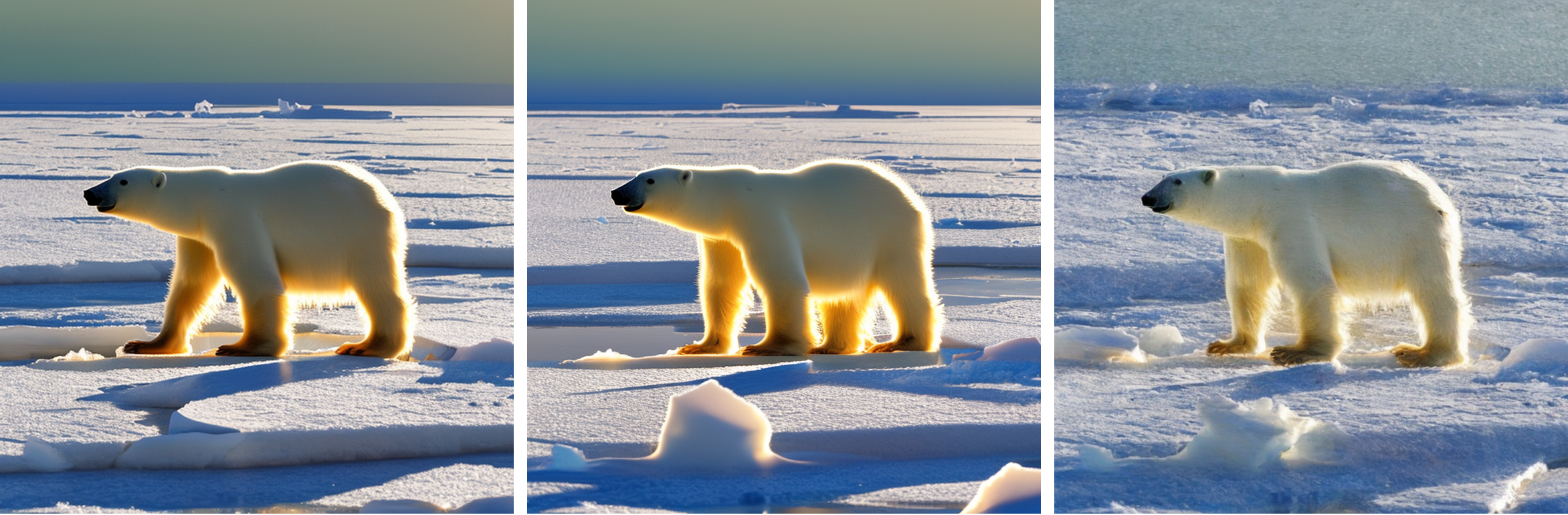}%
\hfill
\includegraphics[width=0.48\linewidth]{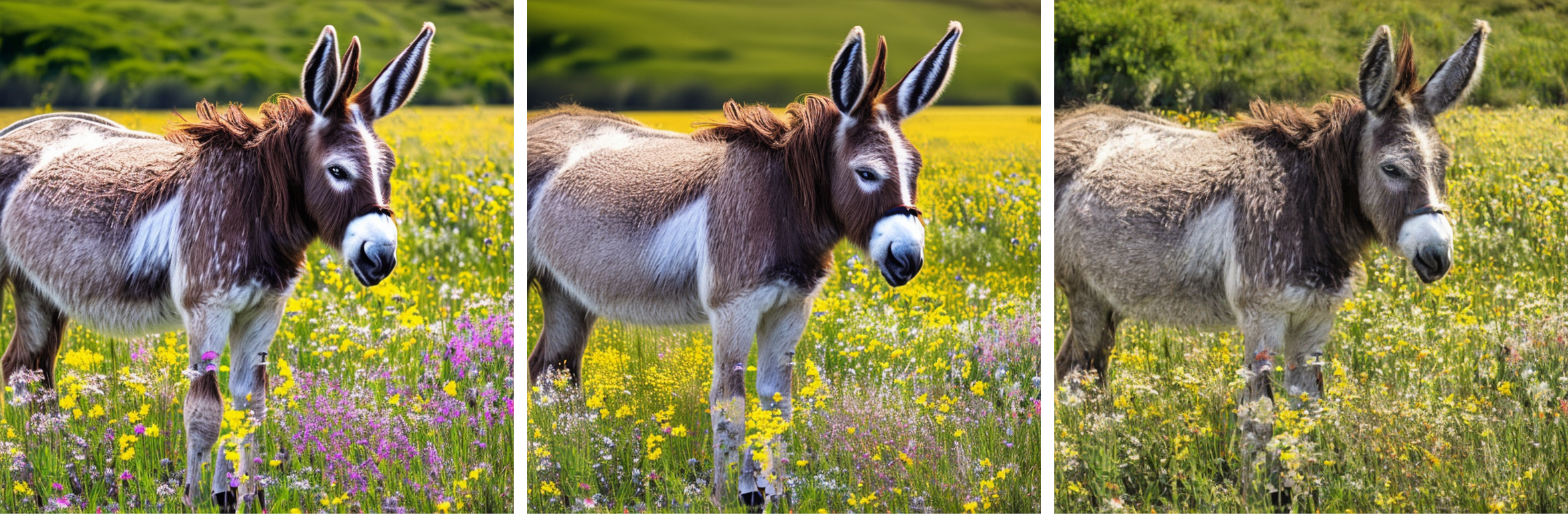}
\\
\includegraphics[width=0.48\linewidth]{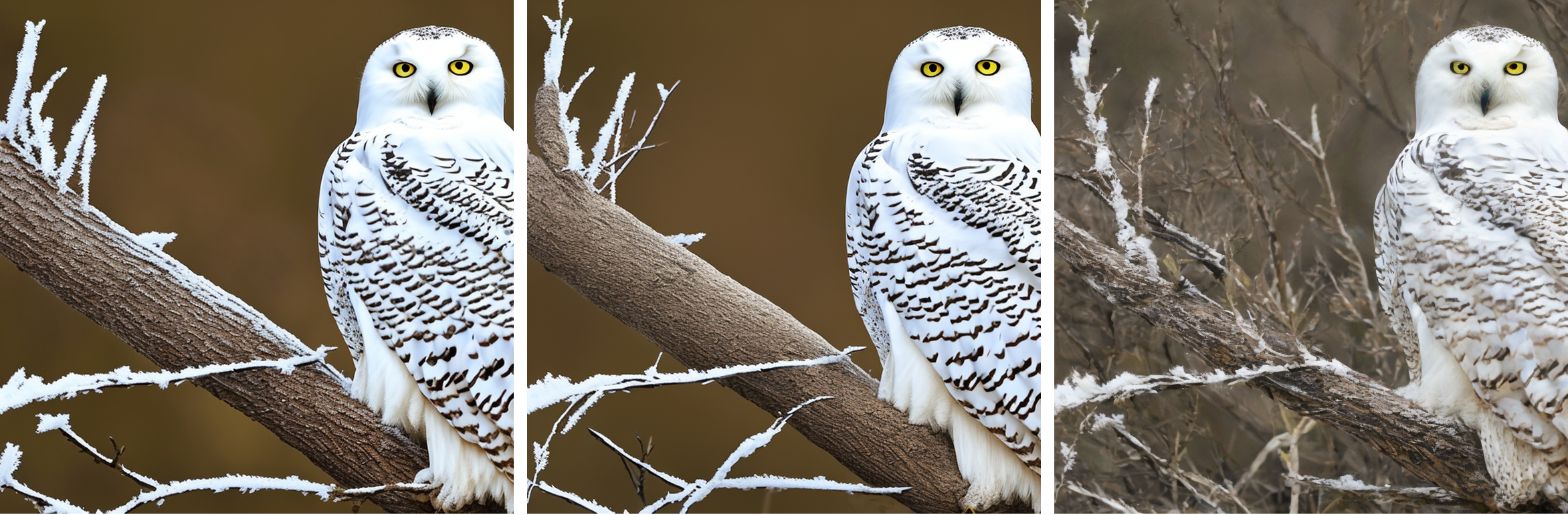}%
\hfill
\includegraphics[width=0.48\linewidth]{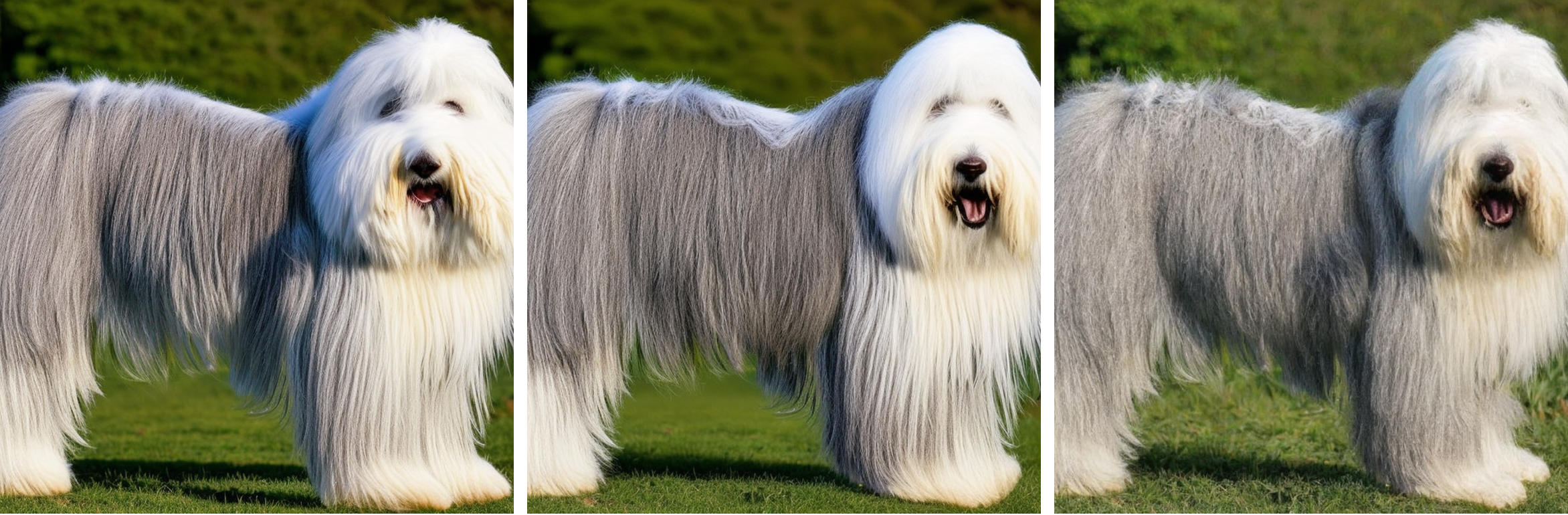}%
\\
\includegraphics[width=0.48\linewidth]{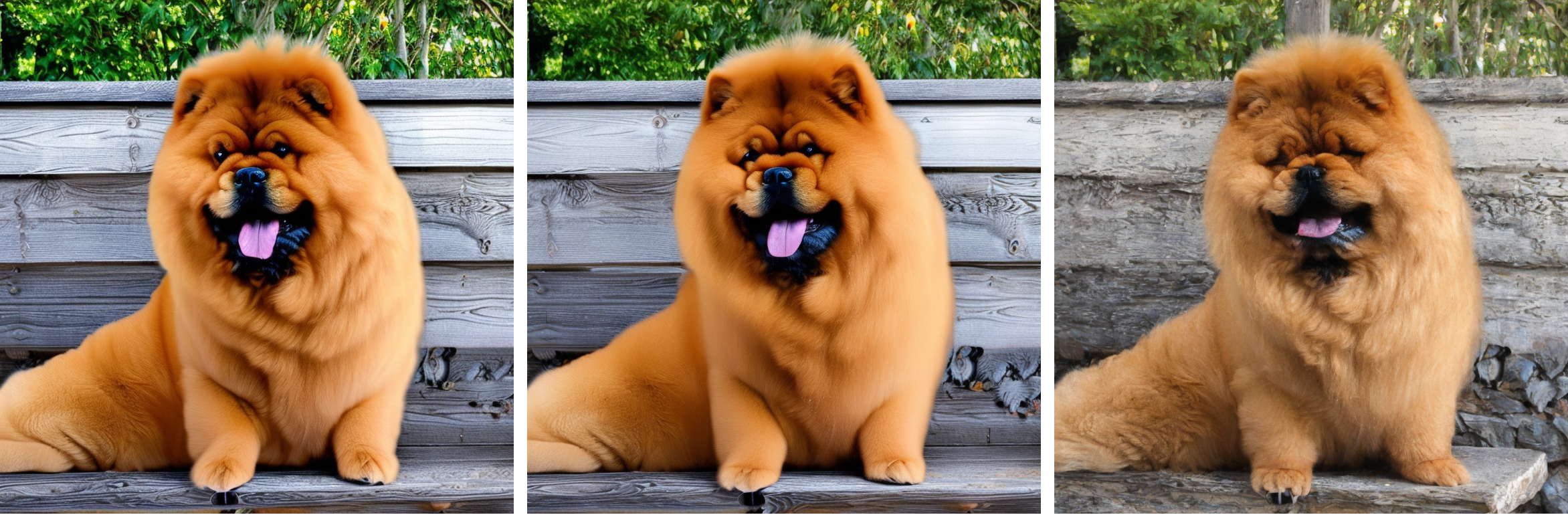}%
\hfill
\includegraphics[width=0.48\linewidth]{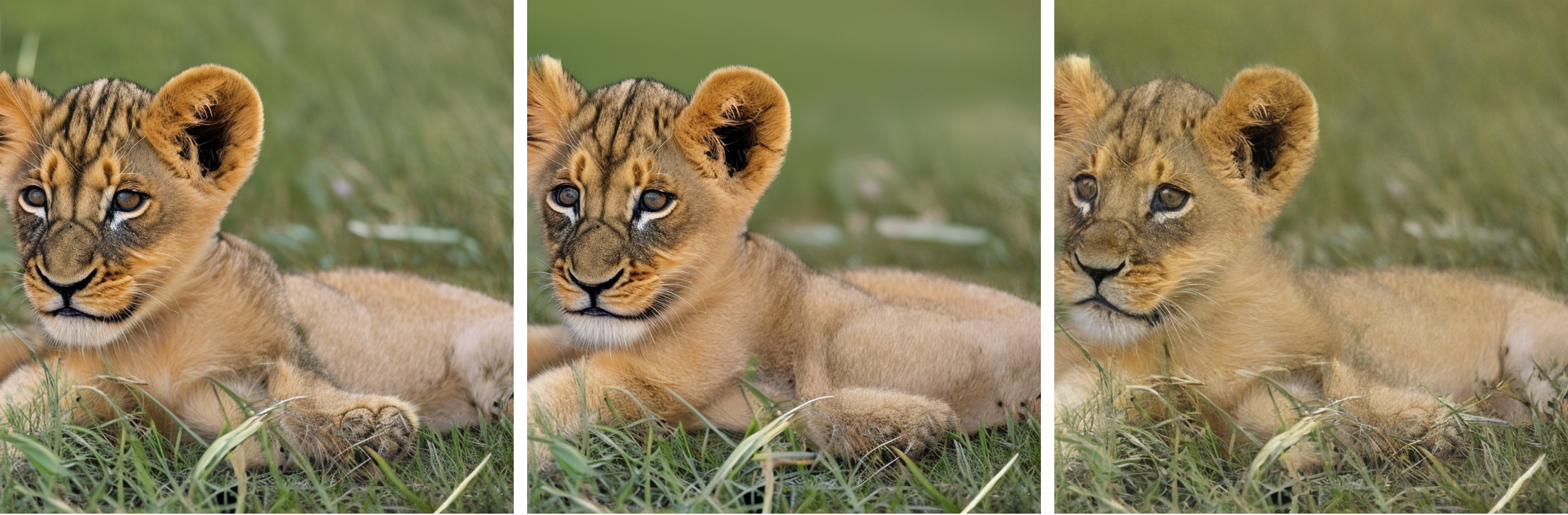}%
\\
\includegraphics[width=0.48\linewidth]{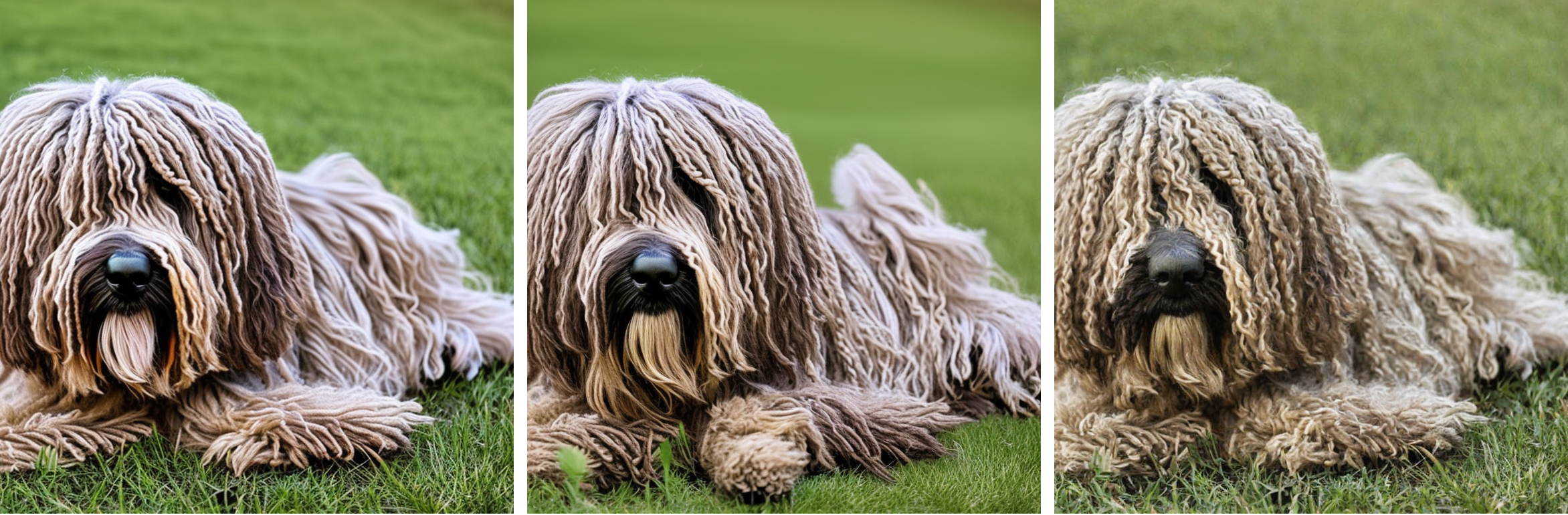}%
\hfill
\includegraphics[width=0.48\linewidth]{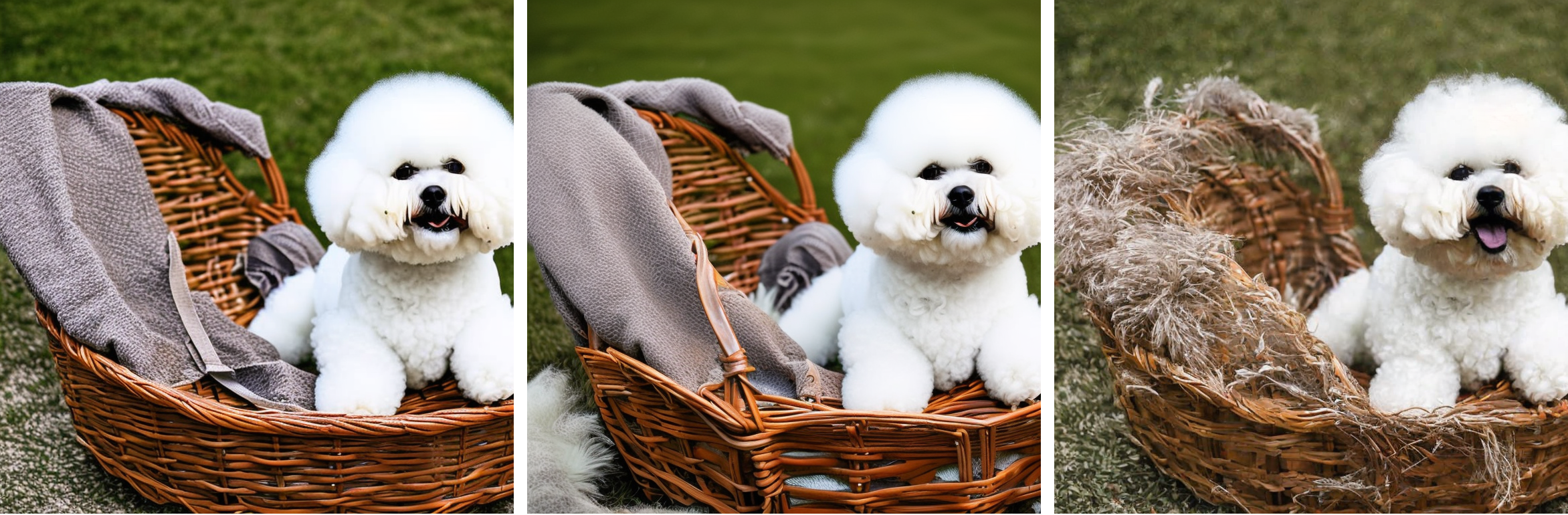}
\\
\includegraphics[width=0.48\linewidth]{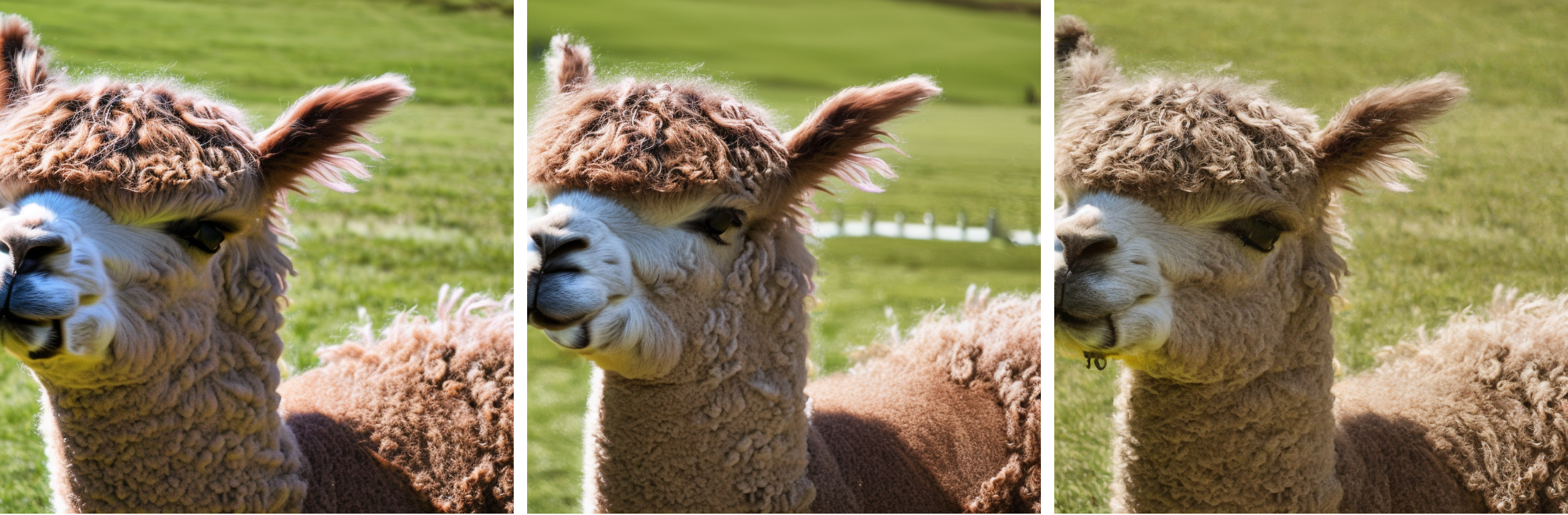}%
\hfill
\includegraphics[width=0.48\linewidth]{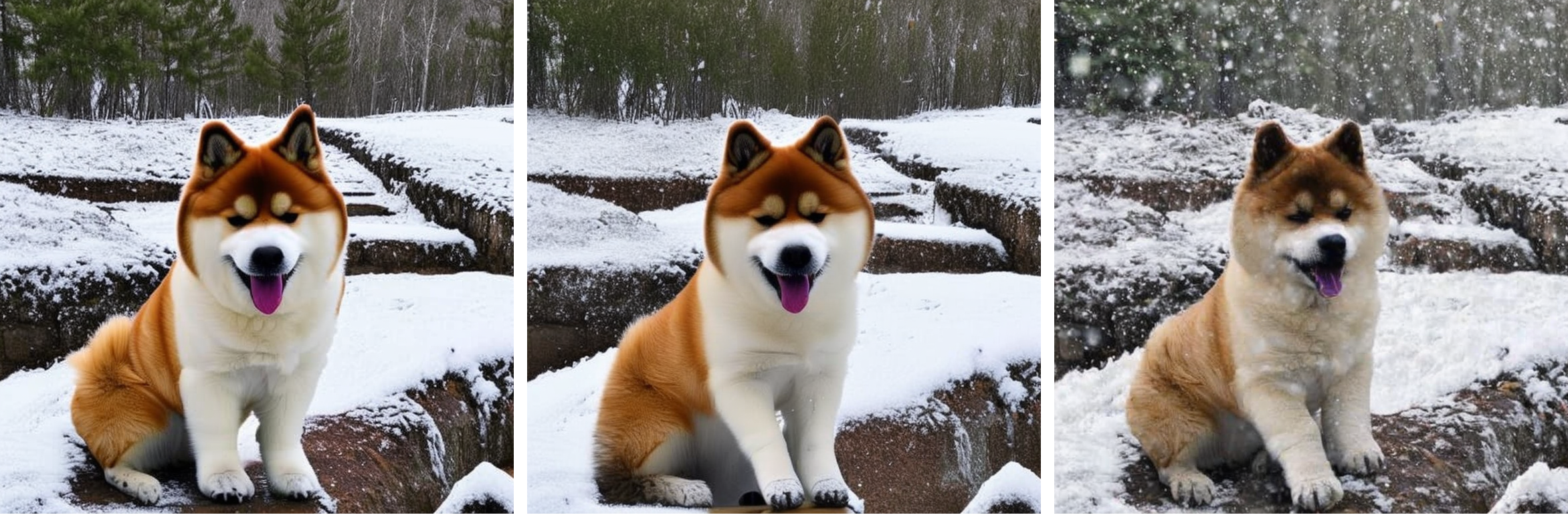}
\\
\includegraphics[width=0.48\linewidth]{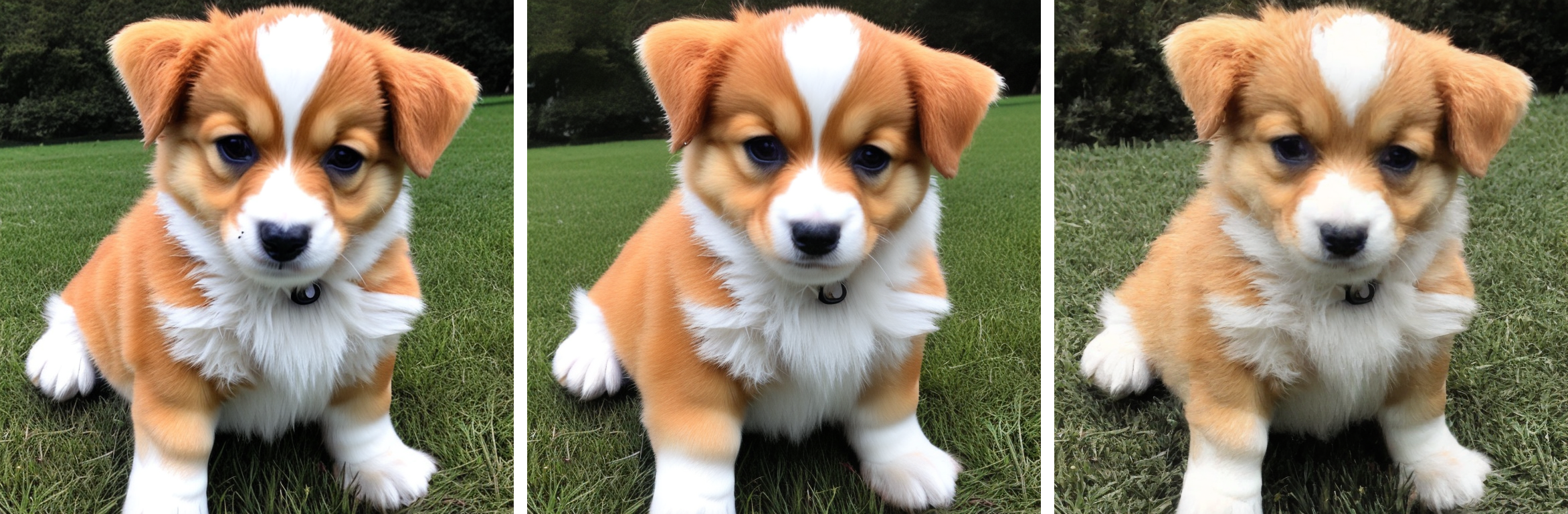}%
\hfill
\includegraphics[width=0.48\linewidth]{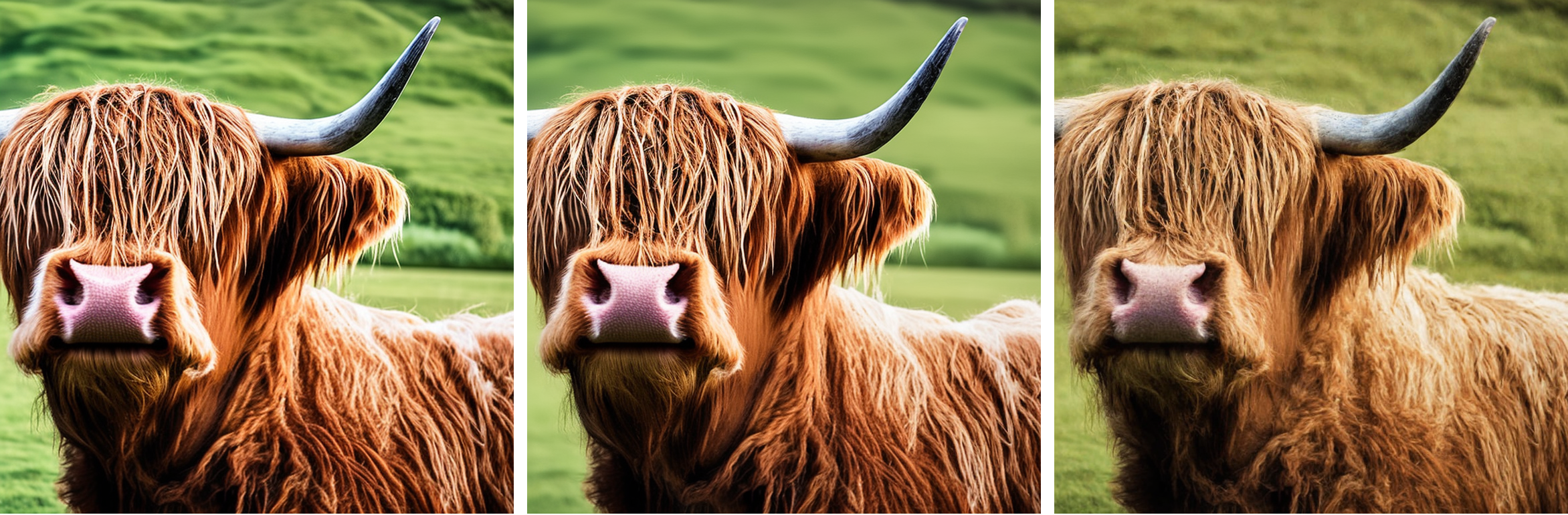}
\\
\includegraphics[width=0.48\linewidth]{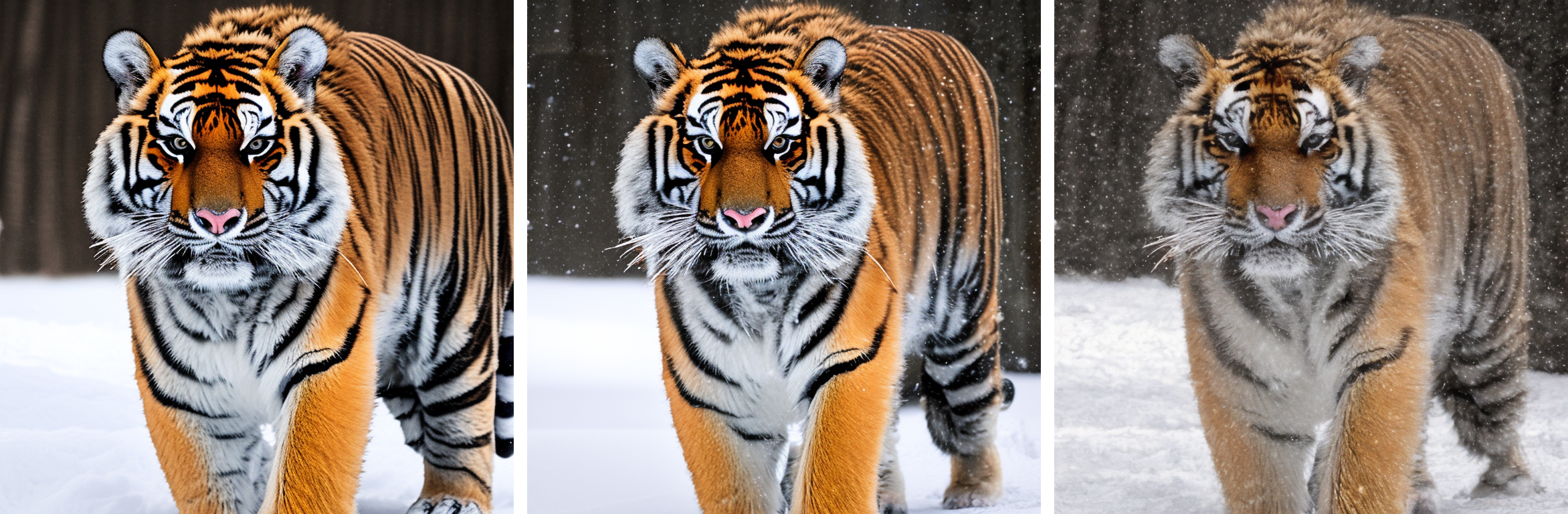}
\hfill
\includegraphics[width=0.48\linewidth]{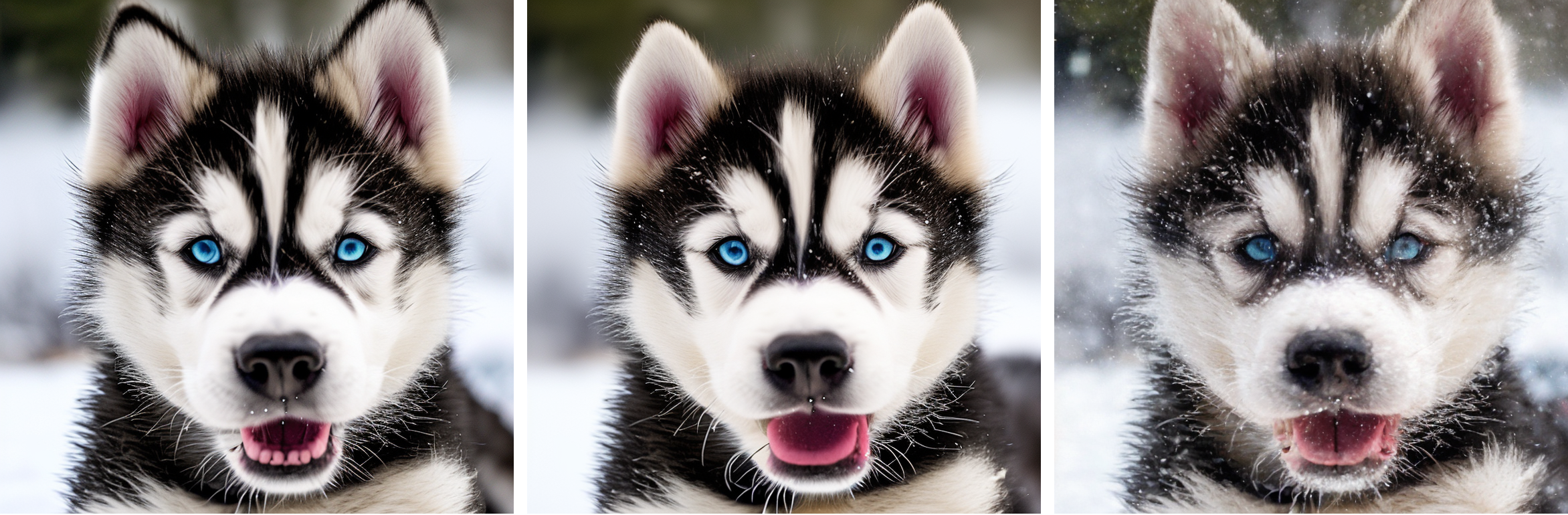}
\\
\includegraphics[width=0.48\linewidth]{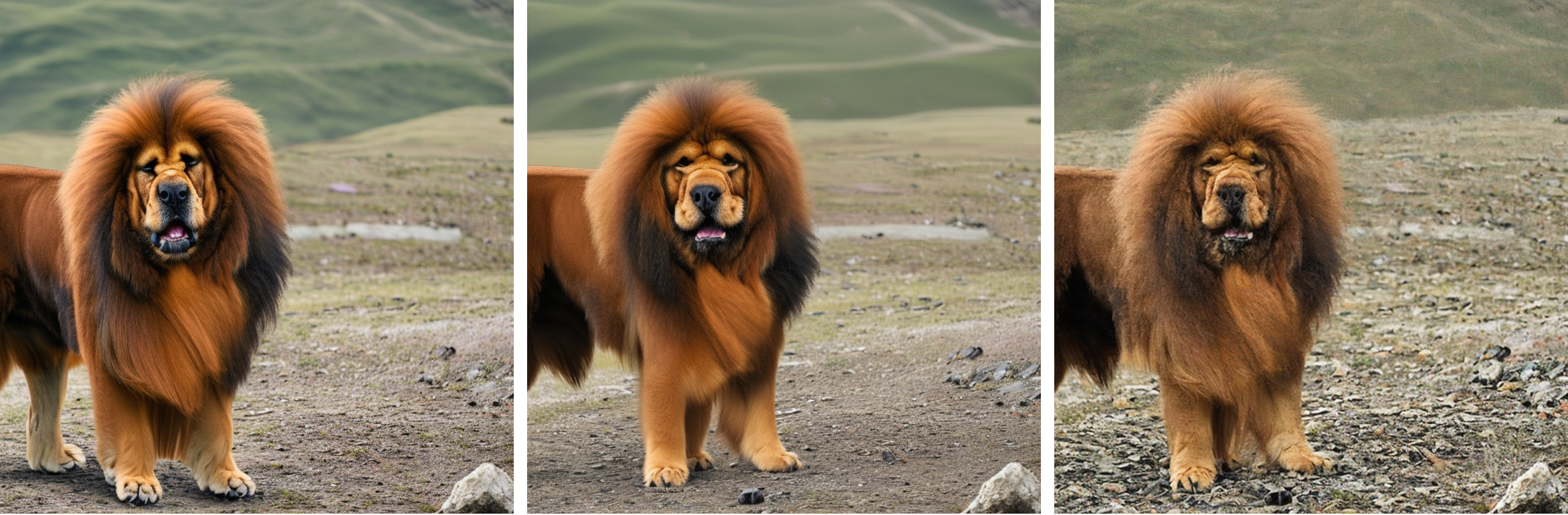}%
\hfill
\includegraphics[width=0.48\linewidth]{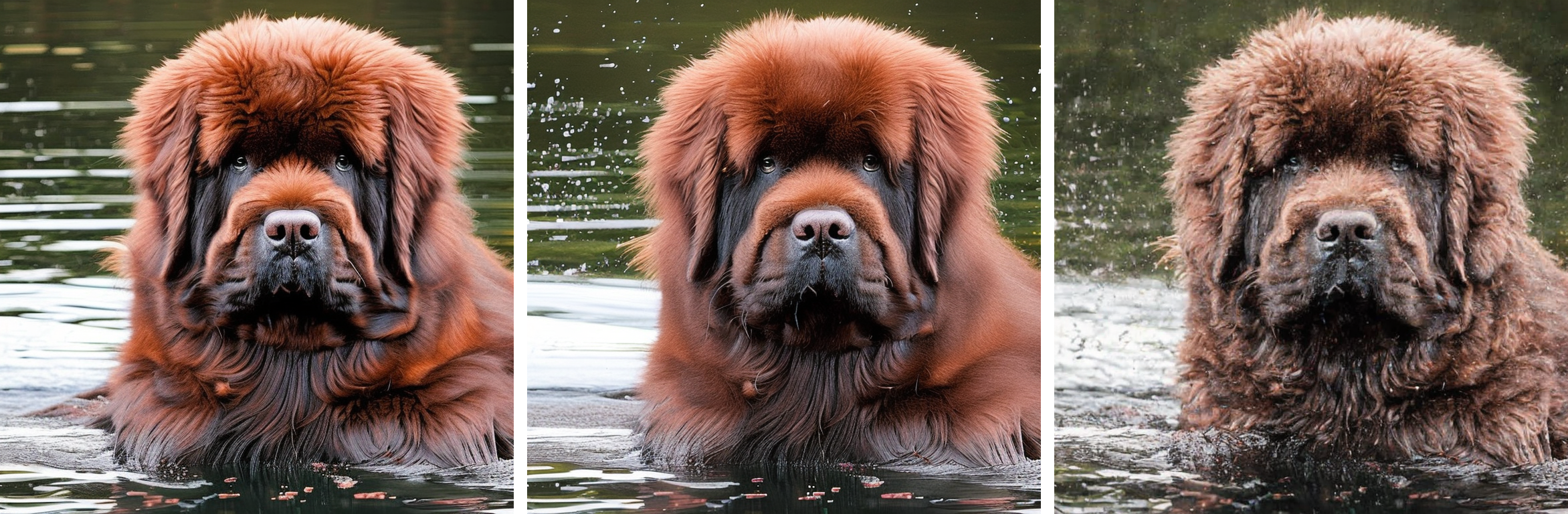}
\vspace{-5pt}
\caption{Images from left to right generated by Stable Diffusion v2.1, ToMeSD and MaMe+MaRe.  MaMe+MaRe retains high-frequency details, making the picture more realistic.}
\label{fig:hair-animals}
\end{figure*}

\begin{figure*}[!ht]
    \centering
    \setlength{\tabcolsep}{0.1pt} 
    \begin{tabular}{cccc}
        \includegraphics[width=0.24\linewidth]{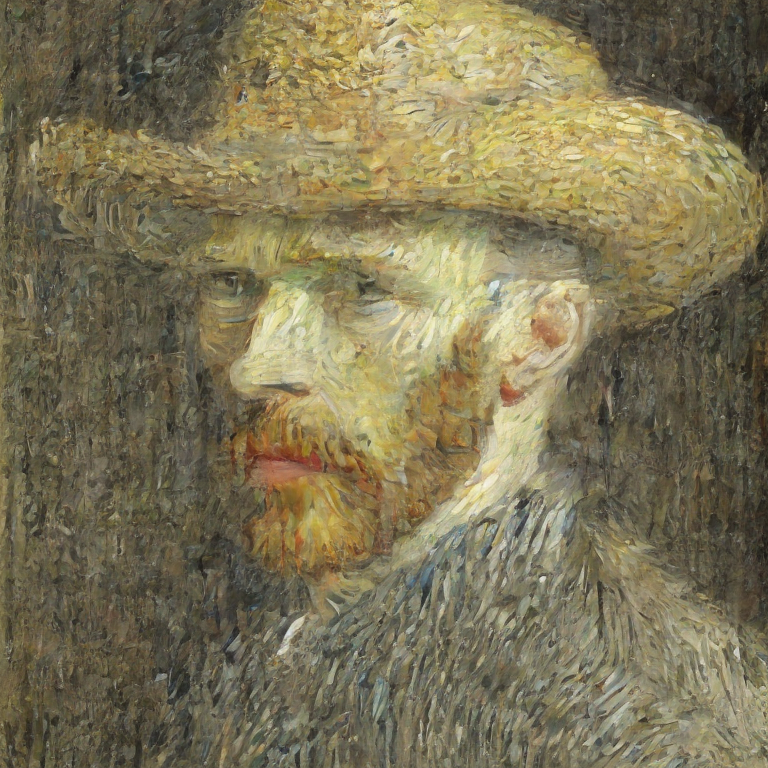} &
        \includegraphics[width=0.24\linewidth]{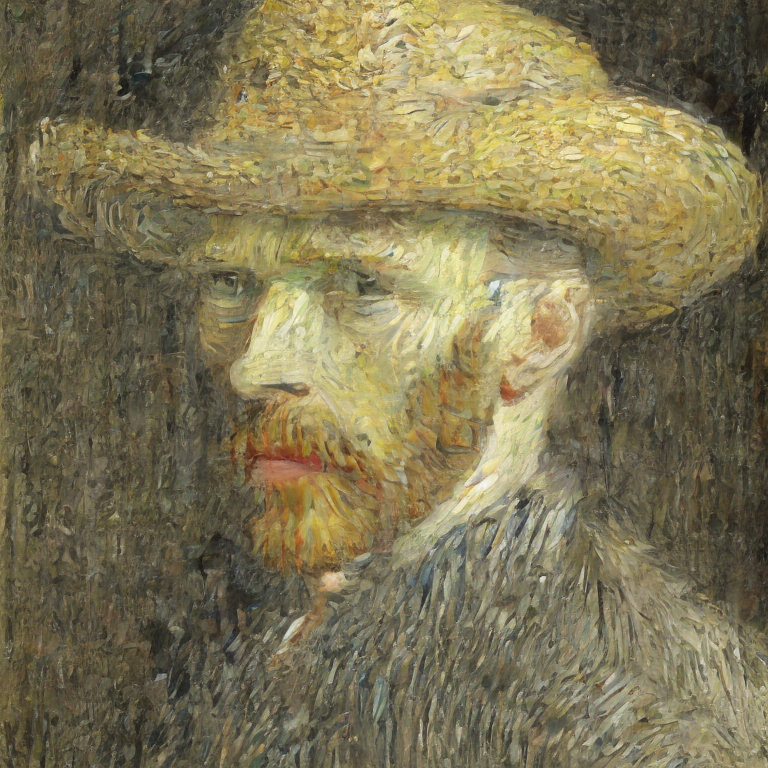} &
        \includegraphics[width=0.24\linewidth]{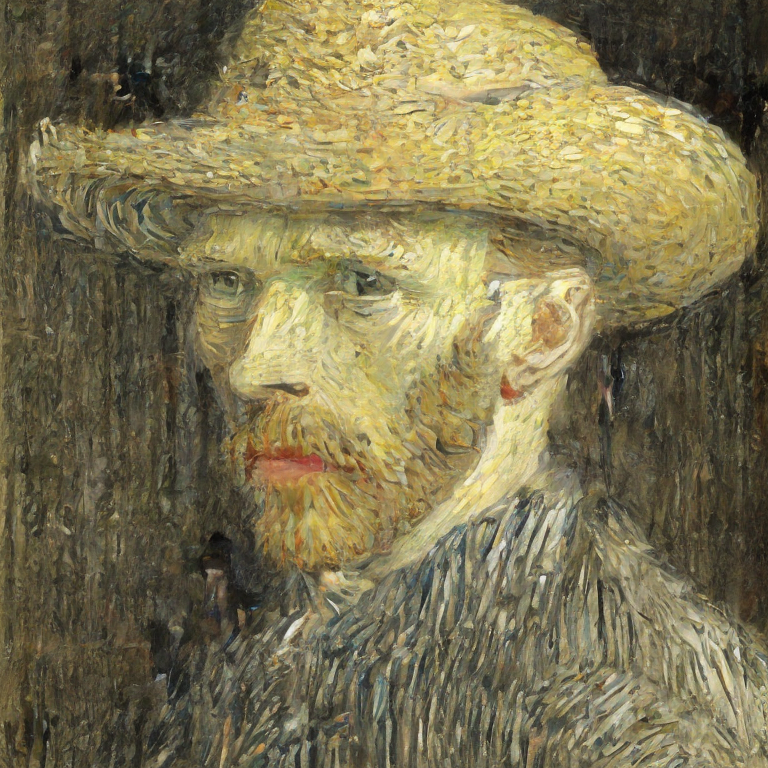} &
        \includegraphics[width=0.24\linewidth]{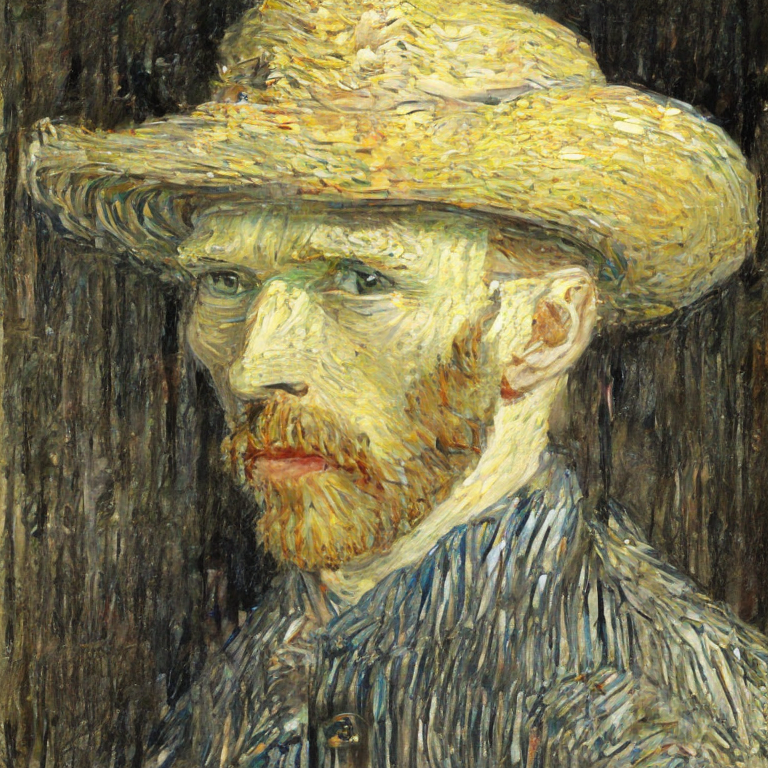} \\

        \includegraphics[width=0.24\linewidth]{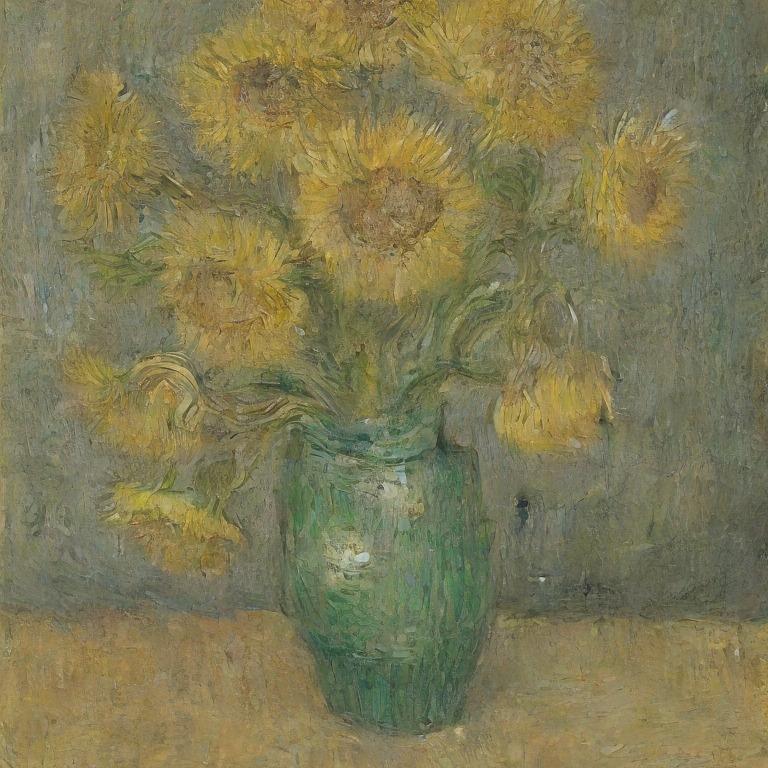} &
        \includegraphics[width=0.24\linewidth]{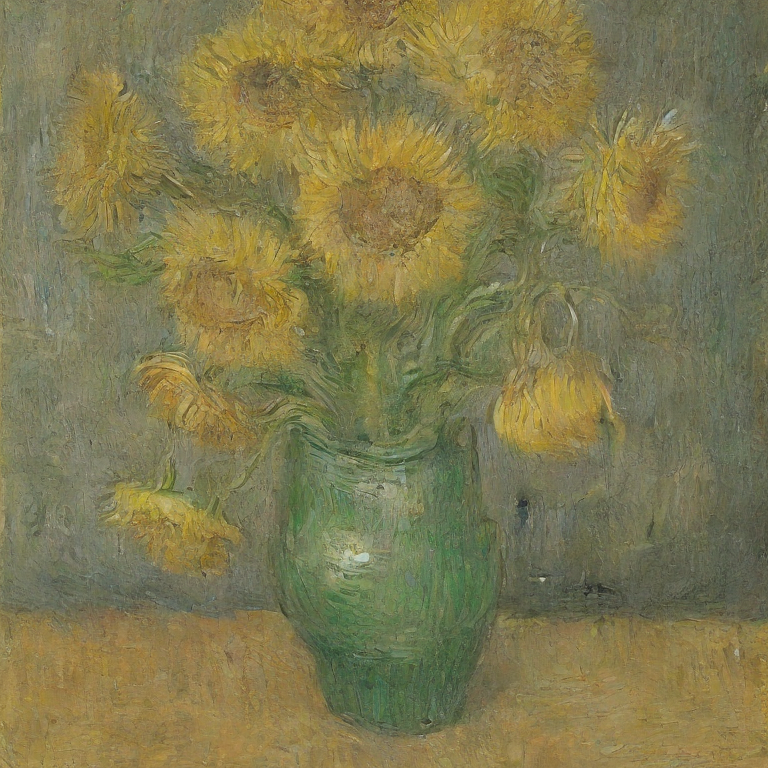} &
        \includegraphics[width=0.24\linewidth]{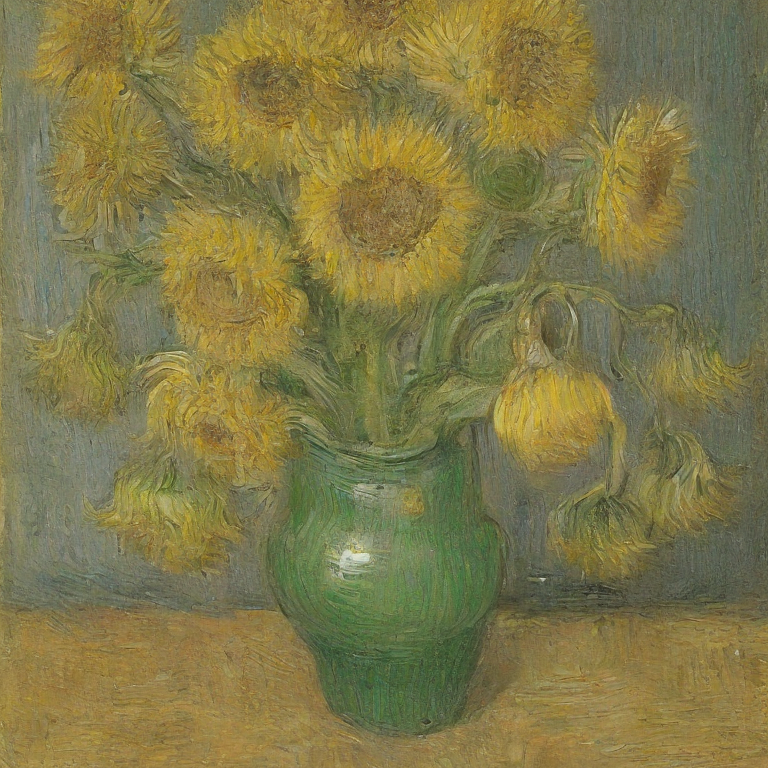} &
        \includegraphics[width=0.24\linewidth]{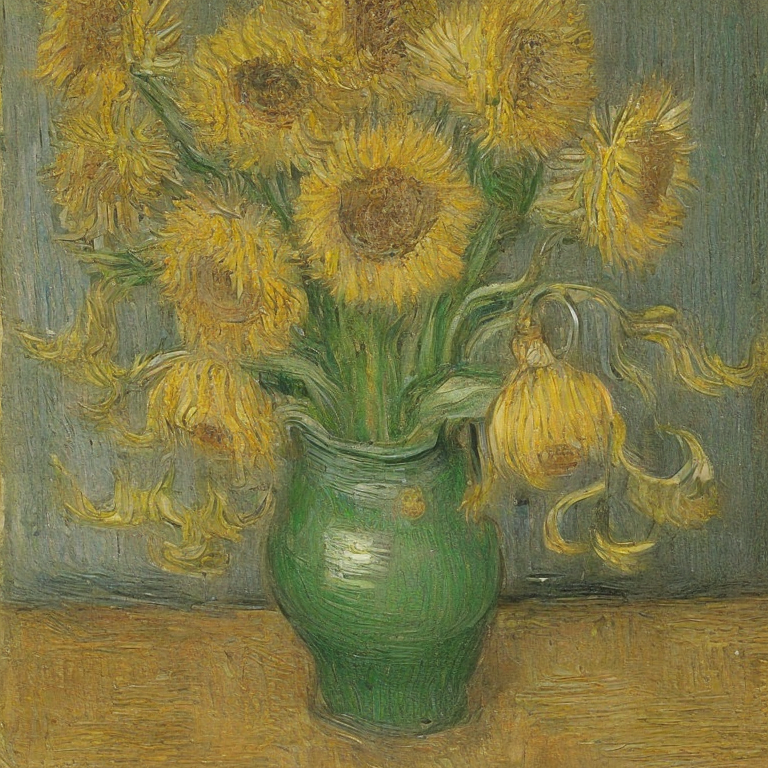} \\

        \includegraphics[width=0.24\linewidth]{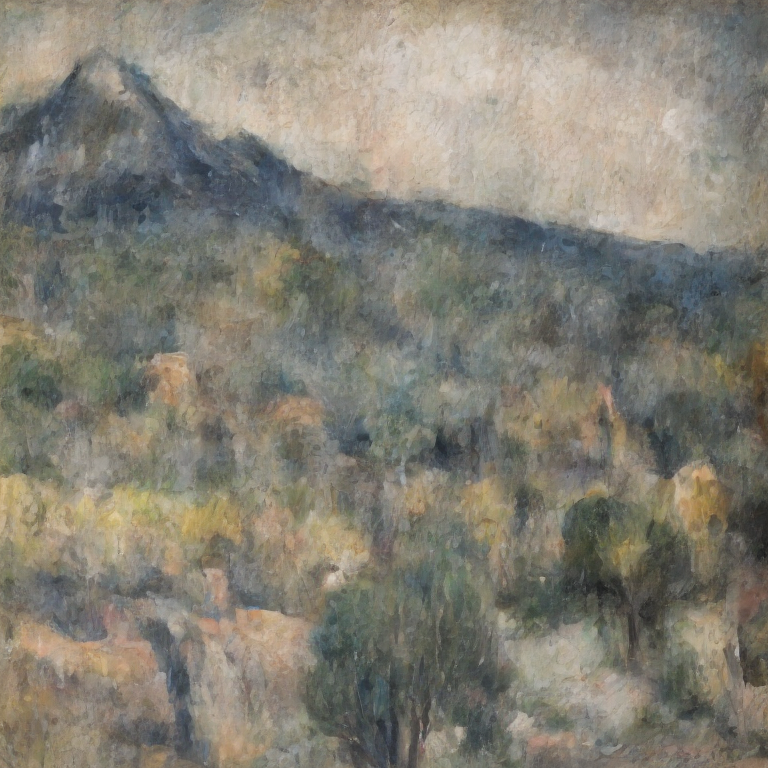} &
        \includegraphics[width=0.24\linewidth]{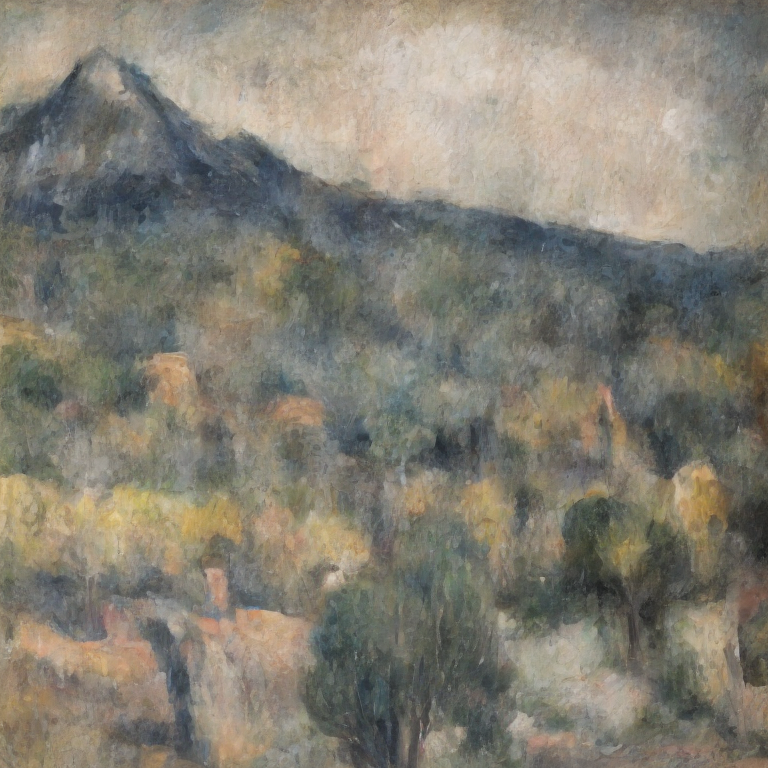} &
        \includegraphics[width=0.24\linewidth]{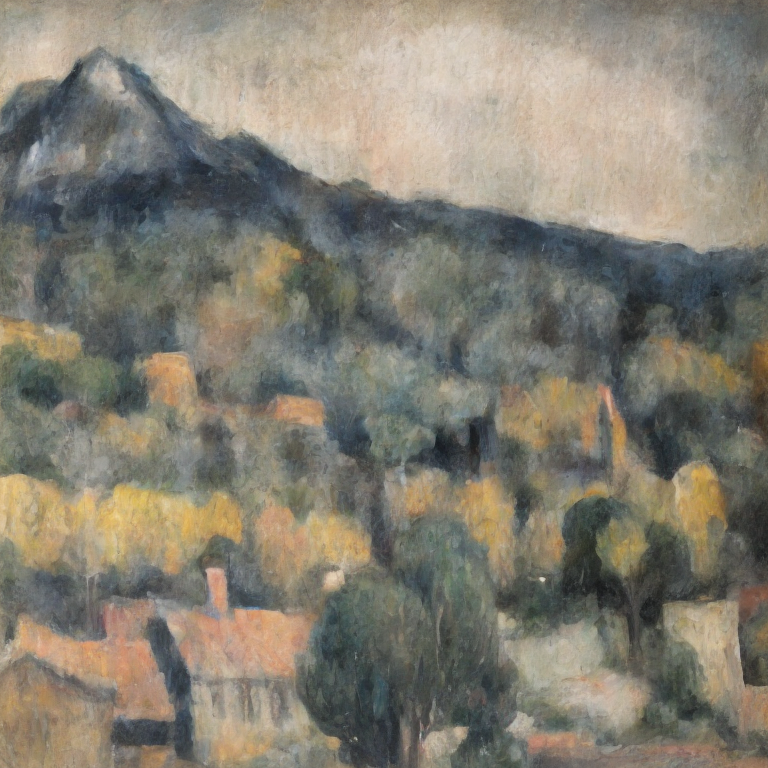} &
        \includegraphics[width=0.24\linewidth]{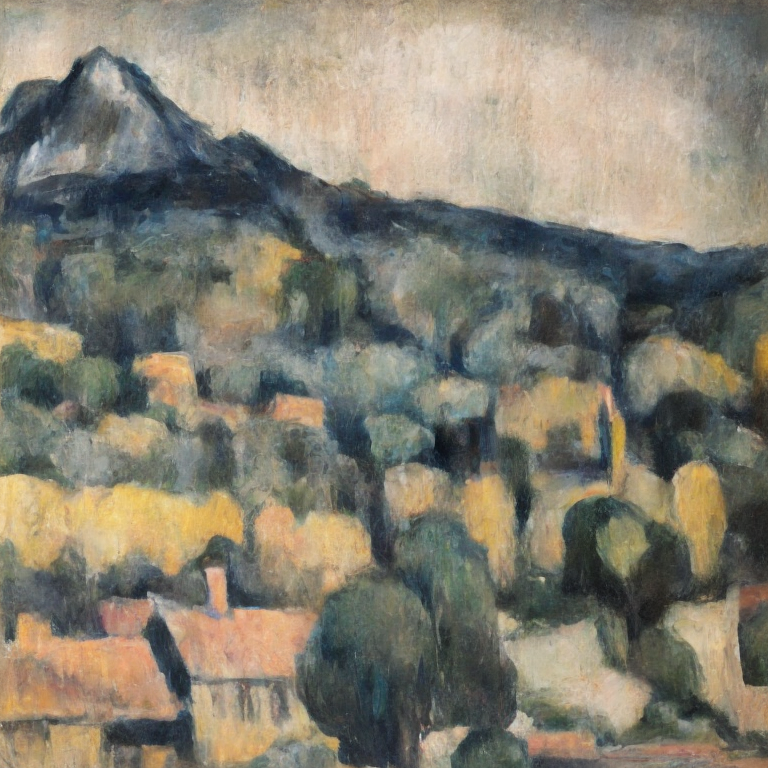} \\

        \includegraphics[width=0.24\linewidth]{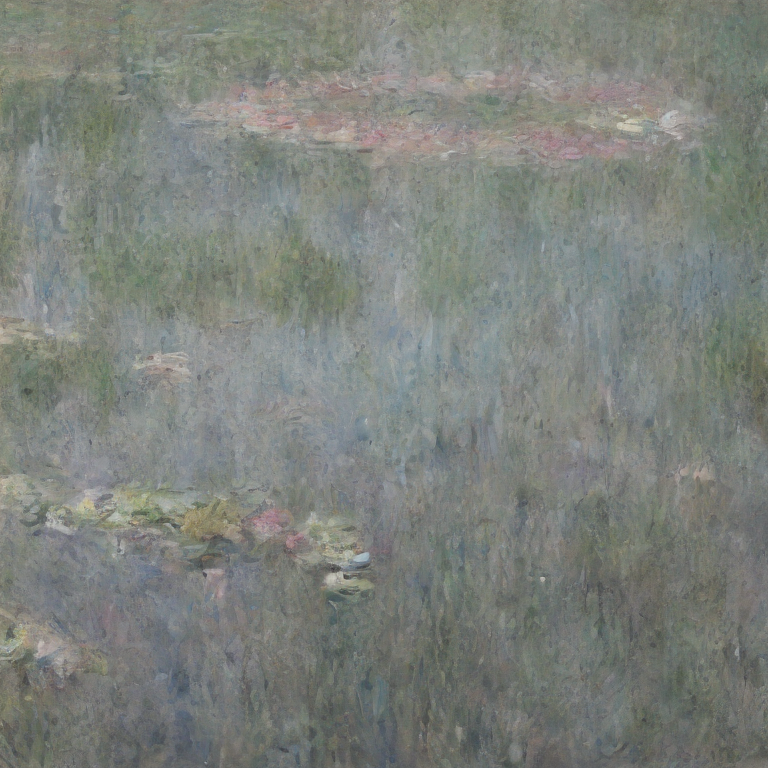} &
        \includegraphics[width=0.24\linewidth]{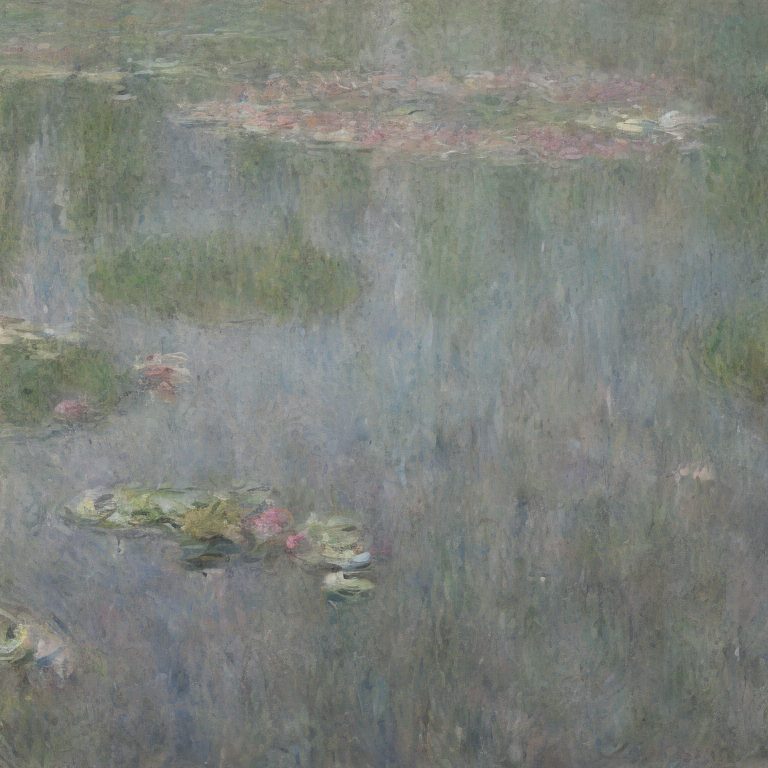} &
        \includegraphics[width=0.24\linewidth]{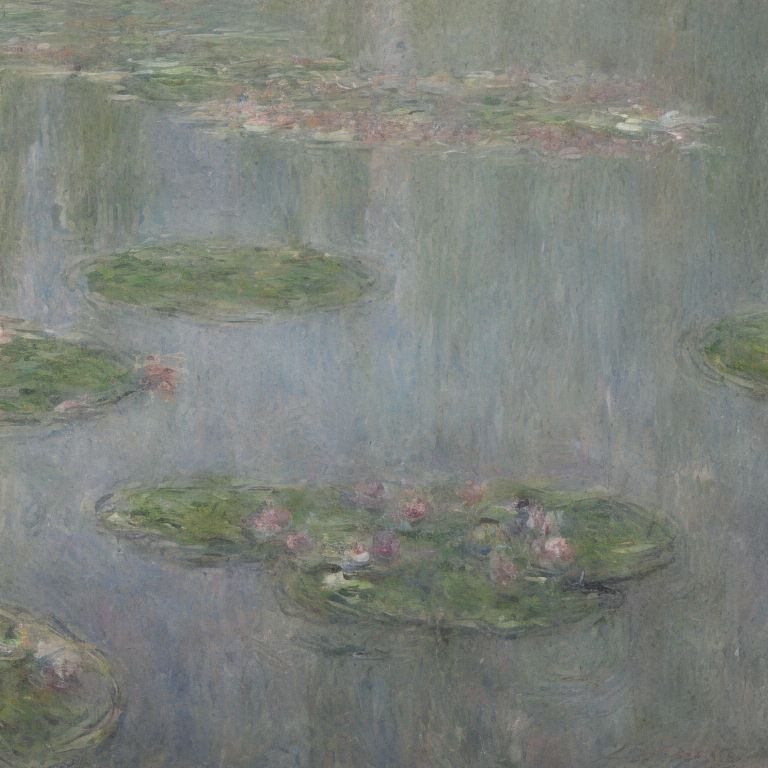} &
        \includegraphics[width=0.24\linewidth]{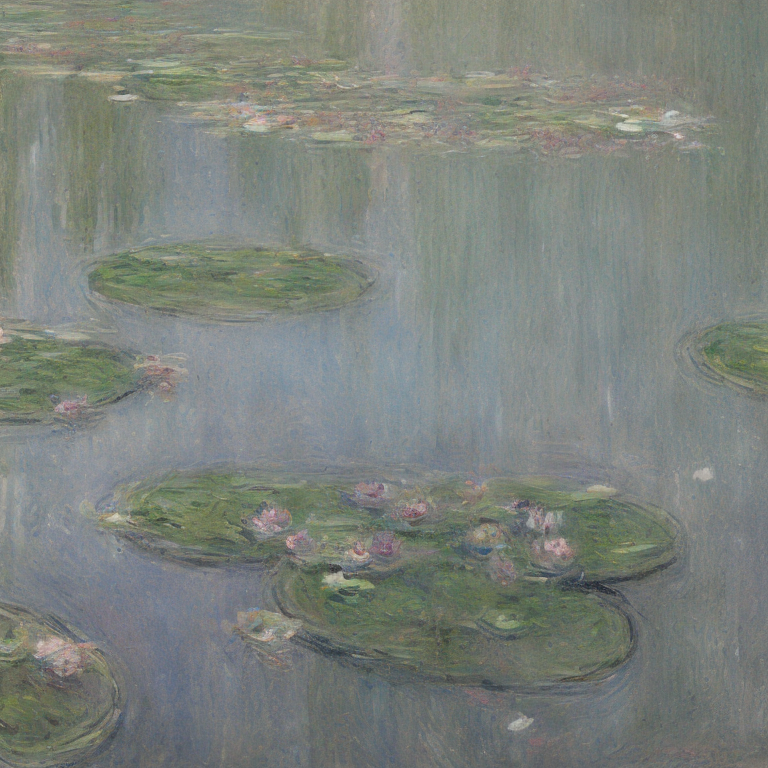} \\

        \includegraphics[width=0.24\linewidth]{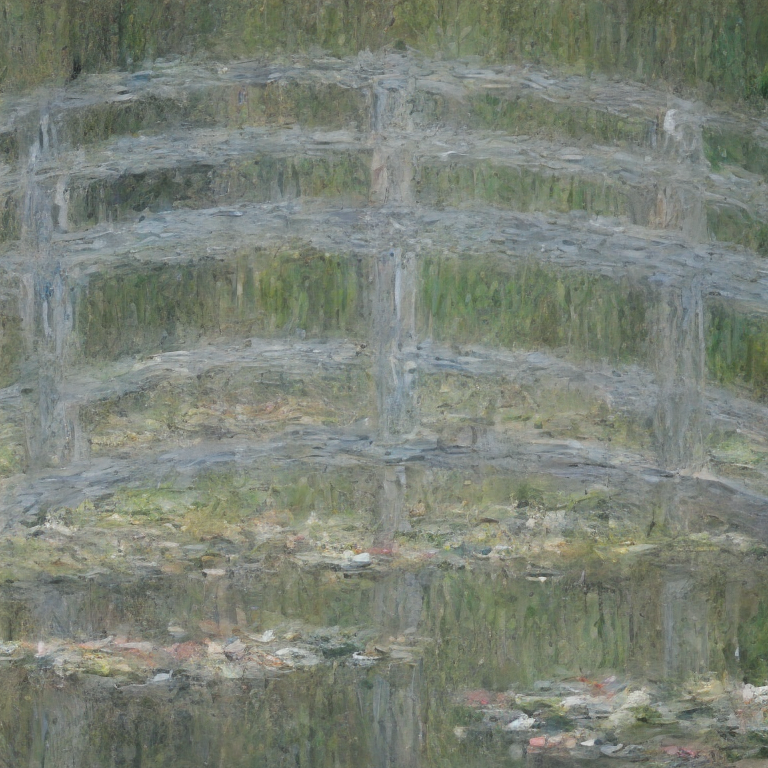} &
        \includegraphics[width=0.24\linewidth]{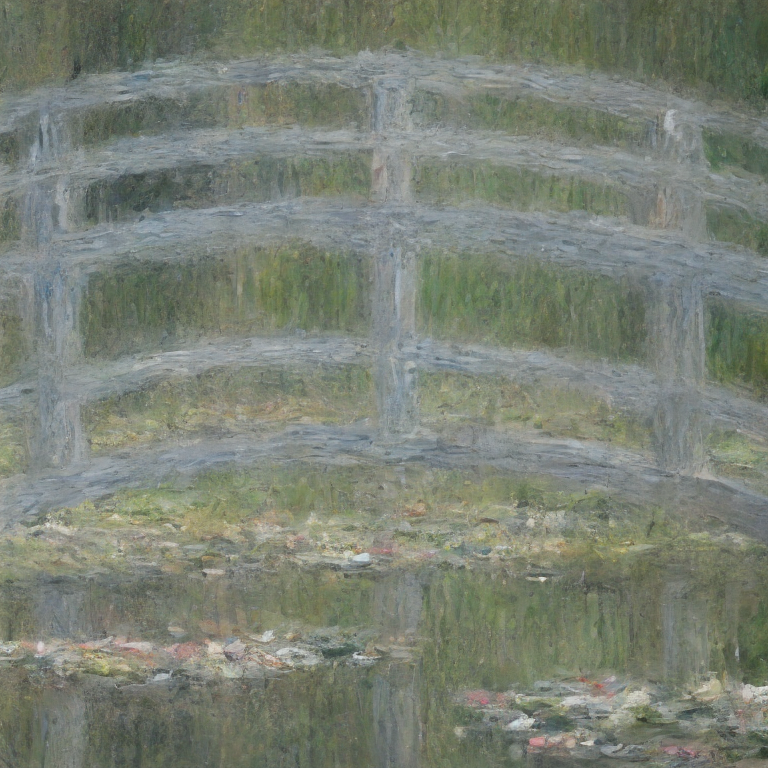} &
        \includegraphics[width=0.24\linewidth]{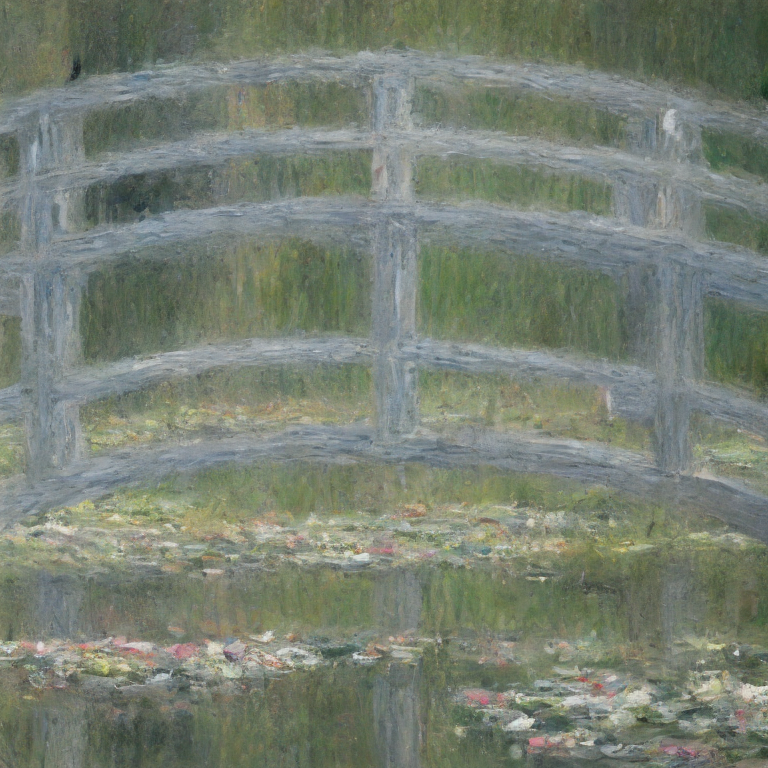} &
        \includegraphics[width=0.24\linewidth]{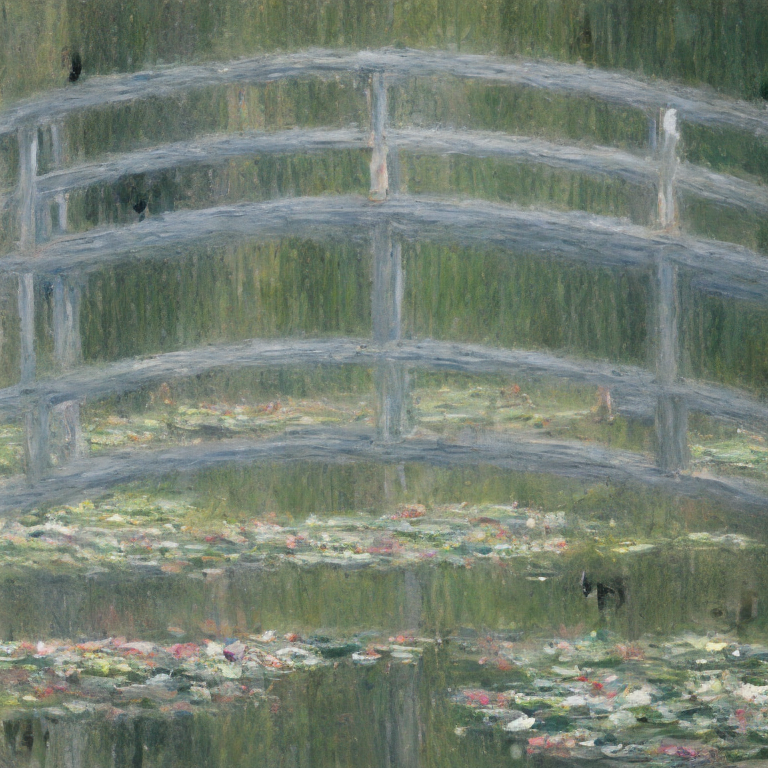} \\

    \end{tabular}
    \vspace{-5pt}
    \caption{Images generated by the model with MaMe+MaRe. The clarity of the painting changed with the similarity threshold. A higher similarity threshold brings a more detailed and clearer painting.}
    \label{fig:imgen_impression}
\end{figure*}

\end{document}